\documentclass[a4paper]{article}

\usepackage{amsmath}
\usepackage{algorithm}
\usepackage[noend]{algpseudocode}
\usepackage{graphicx}
\usepackage{units}
\usepackage{cancel}
\usepackage{epstopdf}
\usepackage{subfigure}
\usepackage{subeqnarray}
\usepackage{soul}

\usepackage{hyperref}
\usepackage{booktabs}
\usepackage{float}
\usepackage[numbers]{natbib}

\usepackage{subfigure}
\makeatletter
\renewcommand{\@thesubfigure}{\normalsize(\textbf{\alph{subfigure}})}
\makeatother

\title{Benchmarking Particle Filter Algorithms for Efficient Velodyne-Based Vehicle Localization}

\author{Jose Luis Blanco-Claraco$^{1}$, Francisco Mañas-Alvarez $^{2}$,\\
Jose Luis Torres-Moreno $^{1,}$*, Francisco~Rodriguez $^{2}$ \\
and Antonio Gimenez-Fernandez $^{1}$}

\date{}

\begin{document}

\maketitle

\begin{abstract}
Keeping a vehicle well-localized within a prebuilt-map is at the core of any autonomous vehicle navigation system. In this work, we show that both standard SIR sampling and rejection-based optimal sampling are suitable for efficient (10 to 20 ms) real-time pose tracking without feature detection that is using raw point clouds from a 3D LiDAR. Motivated by the large amount of information captured by these sensors, we perform a systematic statistical analysis of how many points are actually required to reach an optimal ratio between efficiency and positioning accuracy. Furthermore, initialization from adverse conditions, e.g., poor GPS signal in urban canyons, {we also identify the optimal particle filter settings required to ensure convergence.
Our findings include that a decimation factor between 100 and 200 on incoming point clouds provides a large savings in computational cost with a negligible loss in localization accuracy for a VLP-16 scanner.
Furthermore, an initial density of $\sim$2 particles/m$^2$ is required to achieve 100\% convergence success for large-scale ($\sim$100,000 \text{m}$^2$), outdoor global localization without any additional hint from GPS or magnetic field sensors.}
All implementations have been released as open-source software.
\end{abstract}
	
\section{Introduction}
\label{sec:introduction}
Autonomous vehicles require a robust and efficient localization system capable of
fusing all available information from different sensors and data sources,
such as metric maps or GIS databases.
Metric maps can be automatically built by means of Simultaneous Localization
and Mapping (SLAM) methods onboard the vehicle or retrieved from an external
entity in charge of the critical task of map building.
At present, some companies already have plans to prepare and serve such
map databases suitable for autonomous vehicle navigation,
e.g., Mapper.AI or Mitsubishi's Mobile Mapping System (MMS).
However, at~present, most research groups build their own maps by means of
SLAM methods or, alternatively, using precise real-time kinematic (RTK)-grade global navigation satellite system (GNSS) solutions.
For benchmarking purposes, researchers have access to multiple public datasets including several sensor types in urban environments~\cite{geiger2013vision,Blanco-Claraco2014,Gaspar2018}.

\begin{figure}[H]
	\centering
	\includegraphics[width=1.0\textwidth]{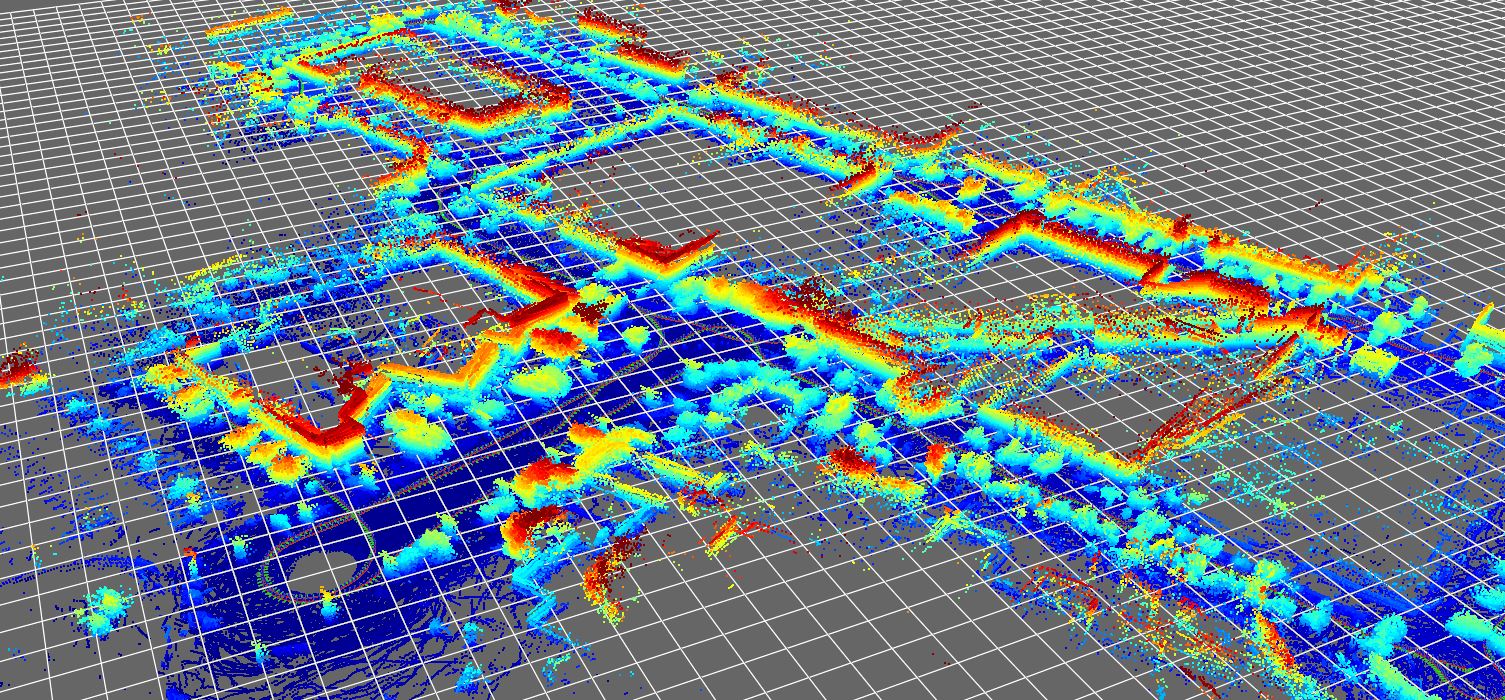}
	\caption{Overview of the ground-truth 3D map used in the benchmarks, representing
		$\sim$100,000 $\text{ m}^2$ of the campus of the University of Almeria, points colorized by height.}
	\label{fig:intro.map.tilted}
\end{figure}

In the present work, we address the suitability of particle filter (PF)
algorithms to localize a vehicle,
equipped with a 3D LiDAR (Velodyne VLP-16, Velodyne Lidar, San Jose, CA, USA),
within~a previously-built reference metric map.
An accurate navigation solution from Novatel (SPAN IGN INS, 
Novatel, Calgary, AB, Canada) is used for RTK-grade
centimeter localization, whose~purpose is twofold:
(i) to help build the reference global map of the environment without the need
to apply any particular SLAM algorithm (see Figure~\ref{fig:intro.map.tilted}),
and (ii) to provide a ground-truth positioning to which the output of the
PF-based localization can be compared in a quantitative~way.

The main contribution of this work is twofold:
(a) providing a systematic
and quantitative evaluation of the trade-off between
how many raw 
points from a 3D-LiDAR must be actually used
in a PF, and~the attainable localization quality;
and (b) benchmarking the particle density that is required
to bootstrap localization, i.e.,~the ``global relocalization'' problem. {For the sake of reproducibility, the~datasets used in this work have been released online (refer to Appendix~\ref{sect:veh.dataset}).}

{The rest of this paper is structured as follows.
	Section~\ref{sec:background} provides an overview of related works in the literature, as~well as some basic mathematical background on
	the employed PF algorithms. 
	Then, Section~\ref{sect:map.sensor.model} proposes an observation model for Velodyne scans, applicable to raw point clouds. 
	{ Next, Section~\ref{sect:justify.decimation} provides mathematical grounds of how a decimation in the input raw LiDAR scan can be understood as an approximation to the underlying likelihood function, and~it is experimentally verified with numerical simulations.}
	The experimental setup is discussed in Section~\ref{sect:experimental}.
	Next, the~results of the benchmarks are presented in Section~\ref{sect:results}
	and we end with some brief conclusions in Section~\ref{sect:conclusions}.}

\section{Background}
\label{sec:background}

{This section provides the required background to put the present proposal in context. 
	We first review related works in Section~\ref{sect:related}, and~
	next, Section~\ref{sect:background.pf} provides further details on the relevant particle filtering algorithms.}

\subsection{Related~Works}
\label{sect:related}

As has been known by the mobile robotics community for more than a decade,
SLAM is arguably more efficiently addressed by means of optimizing
large sparse graphs of observations (some representative works are~\cite{kummerle2011g,ila2017slam++,kaess2012isam2, blanco2019mola})
rather than by means of PF methods.
The latter remain being advantageous only for observation-map pairs whose observation model is neither unimodal nor easy to evaluate in closed form, e.g.,~raw 2D range scans and occupancy grid~maps.

On the other hand, localization on a prebuilt map continues to be a field where PFs find widespread acceptance,
although there are few works in the literature where PFs are applied to the problem of 3D laser range finder (LRF)-based localization of vehicles,
with many works favoring the extraction of features instead of using the raw sensor data.
For example, in~\cite{yoneda2014lidar}, the~authors propose extracting a subset of features (subsets of the raw point cloud)
from Velodyne (HDL-32E) scans, which~are then matched using an iterative closest point algorithm (ICP) against a prebuilt map.
Typical~positioning errors obtained with this method fall in the order of one meter.
Plane features are also extracted by Moosmann and Stiller in~\cite{moosmann2011velodyne} from
Velodyne (HDL-64E) scans to achieve a robust SLAM method.
Interestingly, this work proposes randomly decimating (sampling) the number of features such that
a maximum of 1000 planes are extracted per scan, although~no further details are given regarding
the optimality of this choice.
{Dubé~et~al. proposed a localization framework in~\cite{dube2017segmatch} 
	where features are first extracted from raw 3D LiDAR scans and a descriptor is computed for them. This approach has the advantage of a more compact representation of large/scale maps, enabling~global re-localization faster than with particle filters, although~their computational cost is higher due to the need to segment point clouds, compute descriptors, and~evaluate the matching between them. As~shown in our experimental results, particle filters with decimated 3D LiDAR scans can track a vehicle pose in $\sim$10\, \unit{ms}, whereas~\cite{dube2017segmatch} takes $\sim$800\, \unit{ms}.}

Among the previous proposals to use PFs in vehicle localization and SLAM,
we find~\cite{levinson2007map}, where a PF is used for localizing a vehicle using a reflectance
map of the ground and an associated observation likelihood model.
A modified PF weight-update algorithm is presented in~\cite{rabe2017robust} for precise localization
within lanes by fusing information from visual lane-marking and GPS.
Their method is able to handle probability density functions that mix uniform and Gaussian distributions;
such a flexibility would be also applicable to the optimal-sampling method~\cite{blanco2010ofn}
used in the present work,
which is explained in Section~\ref{sect:background.pf}. The~integration of the localization techniques into vehicle systems' architecture demands more computationally efficient techniques~\cite{Song2018,Maalej2018}. When high level tasks are demanded, as~cooperative driving, real-time performance is critical~\cite{Xu2018}.
Rao-Blackwellized particle filters (RBPF) have been proposed for
Velodyne-based SLAM in~\cite{choi2014hybrid}, using a pre-processing stage where
vertical objects are detected in the raw scans such that RBPF-SLAM
can be fed with a reduced number of discrete features.
The flexibility of observation models in particle filters is exploited
there to fuse information from two different metric maps
simultaneously: a grid-map and the above-mentioned feature map. As~can be seen, PF based methods still remain popular in SLAM. Nowadays, much research in this field focus on reducing their algorithms' computational cost. Thus, in~\cite{Jo2018rbpf}, a~map-sharing strategy is proposed in which the particles only have a small map of the nearby environment.  In~\cite{Vallicrosa2018}, the~Rao-Blackwellized particle filter is executed online by using Hilbert maps. Other fields of interest are the cooperative use of particle filters for multi-robot systems~\cite{Wen2019} and the development of more efficient techniques for the integration of Velodyne sensors~\cite{Grant2018}.

\subsection{Particle Filter~Algorithms}
\label{sect:background.pf}

Particle filtering is a popular name for
a family of sequential Bayesian filtering methods based
on importance sampling~\cite{doucet2001introduction}.
Most commonly, PF in the mobile robot community is used
as a synonymous for the Sequential Importance Sampling (SIS) with resampling (SIR) method,
which is a modification of the Sequential Importance Sampling (SIS) filter
to cope with particle depletion by means of an optional resampling step~\cite{arulampalam2002tpf}.

If we let $\boldsymbol{x}_t$ denote the vehicle pose for time step $t$, the~posterior distribution of the vehicle pose can be computed sequentially by (the full derivation can be found elsewhere~\cite{blanco2009contributions}):
\begin{eqnarray}
\label{optPF:eq:def_pose_post} %
p(\boldsymbol{x}_t|z^t,u^t) \propto
\overbrace{p(z_t| \boldsymbol{x}^t,u^t)}^{\text{Observation likelihood}}
\overbrace{p(\boldsymbol{x}_t|z^{t-1},u^t)}^{\text{Prior}},
\end{eqnarray}

\noindent where $z^t$ and $u^t$ represent the sequences of
robot observations and actions, respectively, for~all past
timesteps up to $t$.
The posterior is approximated by means of a discrete set of $i=1,\dots ,N$
weighted samples $\boldsymbol{x}^{[i]}_{t}$
(called particles)  that represent hypotheses of
the current vehicle pose.
To propagate particles from one timestep to the next,
a particular proposal distribution $q(\boldsymbol{x}_t|\boldsymbol{x}^{t-1,[i]},z^t,u^t)$
must be used, then the weights are updated accordingly by:
\begin{eqnarray}
\label{optPF:eq:SIS_weights} %
\omega_t^{[i]} \propto \omega_{t-1}^{[i]}
\frac{p(z_t|\boldsymbol{x}_t,\boldsymbol{x}^{t-1,[i]},z^{t-1},u^t)p(\boldsymbol{x}_t|\boldsymbol{x}^{[i]}_{t-1},u_t)}{q(\boldsymbol{x}_t|\boldsymbol{x}^{t-1,[i]},z^t,u^t)}.
\end{eqnarray}

Most works in robot and vehicle localization assume that $q(\cdot)$ is the robot motion model $p(\boldsymbol{x}_t|\boldsymbol{x}^{[i]}_{t-1},u_t)$ for convenience,
since, in~that case, Equation~(\ref{optPF:eq:SIS_weights}) simplifies to just evaluating the sensor observation likelihood function $p(z_t|\boldsymbol{x}_t,\cdots)$.
{Following~\cite{blanco2010ofn}, we will refer to this choice as the ``standard proposal'' function.
	Despite its widespread use, it is far from the optimal proposal distribution~\cite{doucet2000sequential}, which by design minimizes the variance of particle weights},
i.e., it maximizes the representativity of particles as samples of the actual distribution being estimated.
Unfortunately, the~optimal solution does not have a closed-form solution
in many practical problems, hence our former proposal of a rejection sampling-based approximation to the optimal PF algorithm in~\cite{blanco2010ofn}.
In~the present work, we will evaluate both PF algorithms, the~``standard'' (SIR with $q(\cdot)$ the motion model from vehicle odometry) and the ``optimal'' proposal distributions (as described in~\cite{blanco2010ofn}), applied~to the problem of vehicle localization.
It is worth highlighting that the latter method is based on the general
formulation of a PF, avoiding the need to perform scan matching (ICP)
between point clouds and hence preventing potential localization failures
in highly dynamic scenarios or in feature-less areas{, 
	where information from other sensors (e.g., odometry) is seamlessly fused
	in the filter leading to a robust localization system.}

{   Our implementation of both PF algorithms features dynamic sample size, using the technique introduced in the seminar work~\cite{fox2003adapting}, to~adapt the computation cost
	to the actual needs depending on how much uncertainty exists at each timestep.
}

\section{Map and Sensor~Model}
\label{sect:map.sensor.model}

A component required by both benchmarked algorithms is the pointwise evaluation of
the sensor likelihood function $p(z_t|x_t,m)$, hence we need to propose one
for Velodyne $360^\circ$ scans ($z_t$) when the robot is at pose $\boldsymbol{x} \in SE(3)$ along a trajectory $\boldsymbol{x}(t)$ given a prebuilt map $m$.

Regarding the metric map $m$, we will assume that it is represented as a 3D point cloud.
We~employed a Novatel's inertial RTK-grade GNSS solution to build the maps for benchmarking and also to obtain the ground-truth vehicle path to evaluate the PF output.
This solution provides us with accurate WGS84 geodetic coordinates, {   as well as heading, pitch and roll attitude angles}.
Using an arbitrary nearby geodetic coordinate as a reference point, coordinates are then converted to a local ENU (East-North-Up) Cartesian frame of reference.
Time interpolation of $\boldsymbol{x}(t)$ is used to estimate the ground-truth path of the Velodyne scanner and the orientation of each laser LED as they rotate to scan the environment; this is known as de-skewing \cite{moosmann2011velodyne} and becomes increasingly important as vehicle dynamics become faster.
{   From each such interpolated pose, we compute the local Euclidean coordinates of the point corresponding to each laser-measured range, then~project it from the interpolated sensor pose in global coordinates. Repeating this for each measured range over the entire data set leads to the generation of the global point cloud of the campus employed as ground-truth map in this work.}


Once a global map is built for reference, we evaluate the likelihood
function $p(z_t|x_t,m)$ as depicted in Algorithm \ref{alg1}.
First, it is worth mentioning the need to work with log-likelihood values when working with a particle filter to extend the valid range of likelihood values that can be represented within machine precision.
The inputs of the observation likelihood (line 1) are the robot pose $\boldsymbol{x}(t)$, a~decimation parameter, the~list of all $N$ points
$\mathbf{p}_l^i$ in local coordinates with respect to the scanner,
the reference map as a point cloud, a~scaling $\sigma$ value that determines how sharp the likelihood function is, and~a smoothing parameter $d_{max}$ that prevent underflowing. Put in words, from~a decimated list of points, each point is first projected to the map coordinate frame (line 6), and~the nearest neighbor is searched for within all map points using a K-Dimensional tree (KD-tree) (line 7).
Next, the~distance between each such local point and its candidate match in the global map is clipped to a maximum $d_{max}$ and the squared distances accumulated into $d^2$. Finally, the~log likelihood is simply $-d^2/\sigma^2$, which implies that we are assuming a truncated (via $d\_max$) Gaussian error model as a likelihood function. Obviously, the~decimation parameter linearly scales the computational cost of the method: larger decimation values provide faster computation speed at the price of discarding potentially valuable information. A~quantitative experimental determination of an optimal value for this decimation is presented later~on.

\begin{algorithm}[H]
	\caption{Observation~likelihood.}
	\label{alg1}
	\hspace*{0.02in} {Input:}
	$\boldsymbol{x}$, decim, $\{\mathbf{p}_l^i\}_{i=1}^{i=N}$, $\mathbf{map}$, $\sigma$, $d_{max}$\\
	\hspace*{0.02in} {Output:}
	loglik
	\begin{algorithmic}[1]
		\setlength\baselineskip{18pt}
		\State begin
		\State		loglik $\gets$ 0
		\State		$d^2$ $\gets$ 0
		\State		foreach i in 1:decim:N
		\State		$\mathbf{p}_g^i \gets \boldsymbol{x} \oplus \mathbf{p}_l^i$
		\State		$\mathbf{p}_{map~closest}^{i}$ $\gets$ map.kdtree.query($\mathbf{p}_g^i$)
		\State		$d^2$ $\gets$ $d^2 + \min\{ d^2_{max}, ||\mathbf{p}_g^i - \mathbf{p}_{map~closest}^{i}||^2  \}$
		\State		end
		\State		loglik $\gets$ $-d^2/\sigma^2$
		\State		return loglik
		\State		end
	\end{algorithmic}
\end{algorithm}

{  
	It is noteworthy that the proposed likelihood model in Algorithm \ref{alg1}
	can be shown to be equivalent to a particular kind of robustified least-square kernel function in the framework of M-estimation~\cite{huber1992robust}. 
	In particular, our cost function is equivalent to the so-called 
	truncated least squares~\cite{audibert2011robust,Carlone-RSS-19}, or~trimmed-mean M-estimator~\cite{ruppert1980trimmed,chen2004m}. 
	
	With $\boldsymbol{x} \in SE(3)$ the vehicle pose to be estimated, a~least-squares formulation to find the optimal pose $\boldsymbol{x}^\star$ that minimizes the 
	total square error between $N$ observed points and their closest correspondences in the map reads:
	\begin{eqnarray}
	\label{eq:ls}
	\boldsymbol{x}^\star &=& \arg\min_\mathbf{x} \sum_{i=1}^{N} c^2_i(\mathbf{x}),  \\
	\text{with:} \quad c_i(\mathbf{x}) &=& \left|\left| \left( \mathbf{x} \oplus \mathbf{p}_l^i \right) - \mathbf{p}_{map~closest}^{i}  \right|\right|.
	\end{eqnarray}
	
	However, this naive application of least-squares suffers from a lack of robustness against outliers: it is well known that a single outlier ruins a least-squares estimator~\cite{ruppert1980trimmed}.
	Therefore, robust M-estimators are preferred, where Equation~(\ref{eq:ls})
	is replaced by:
	\begin{eqnarray}
	\label{eq:robust.ls}
	\boldsymbol{x}^\star &=& \arg\min_\mathbf{x} \sum_{i=1}^{N} f(c_i(\boldsymbol{x}))  
	\end{eqnarray}
	
	\noindent with some robust kernel function $f(c)$. Regular least-squares correspond to the choice \mbox{$f(c)=c^2$}, while~other popular robust cost functions are the Huber loss function~\cite{huber1992robust} or the truncated least-squares~function:
	\begin{eqnarray}
	\label{eq:trunc.ls}
	f(c)  &=& \left\{  
	\begin{array}{ccl}
	c^2  &\quad&  |c|<\theta, \\
	\theta^2  &\quad&  |c| \ge \theta,
	\end{array}
	\right.
	\end{eqnarray}
	
	\noindent which is illustrated in Figure~\ref{fig:robust.ls}. 
	The parameter $\theta$ establishes a threshold for what should be
	considered an outlier. The~insight behind M-estimators is that, 
	by reducing the error assigned to outliers in comparison to a pure least-squares formulation, the~optimizer will tend to ignore them and ``focus'' on minimizing the error of inliers instead that is of those observed points that actually do correspond to map~points.
	
	\begin{figure}[H]
		\centering
		\subfigure[]{
			\includegraphics[width=0.45\columnwidth]{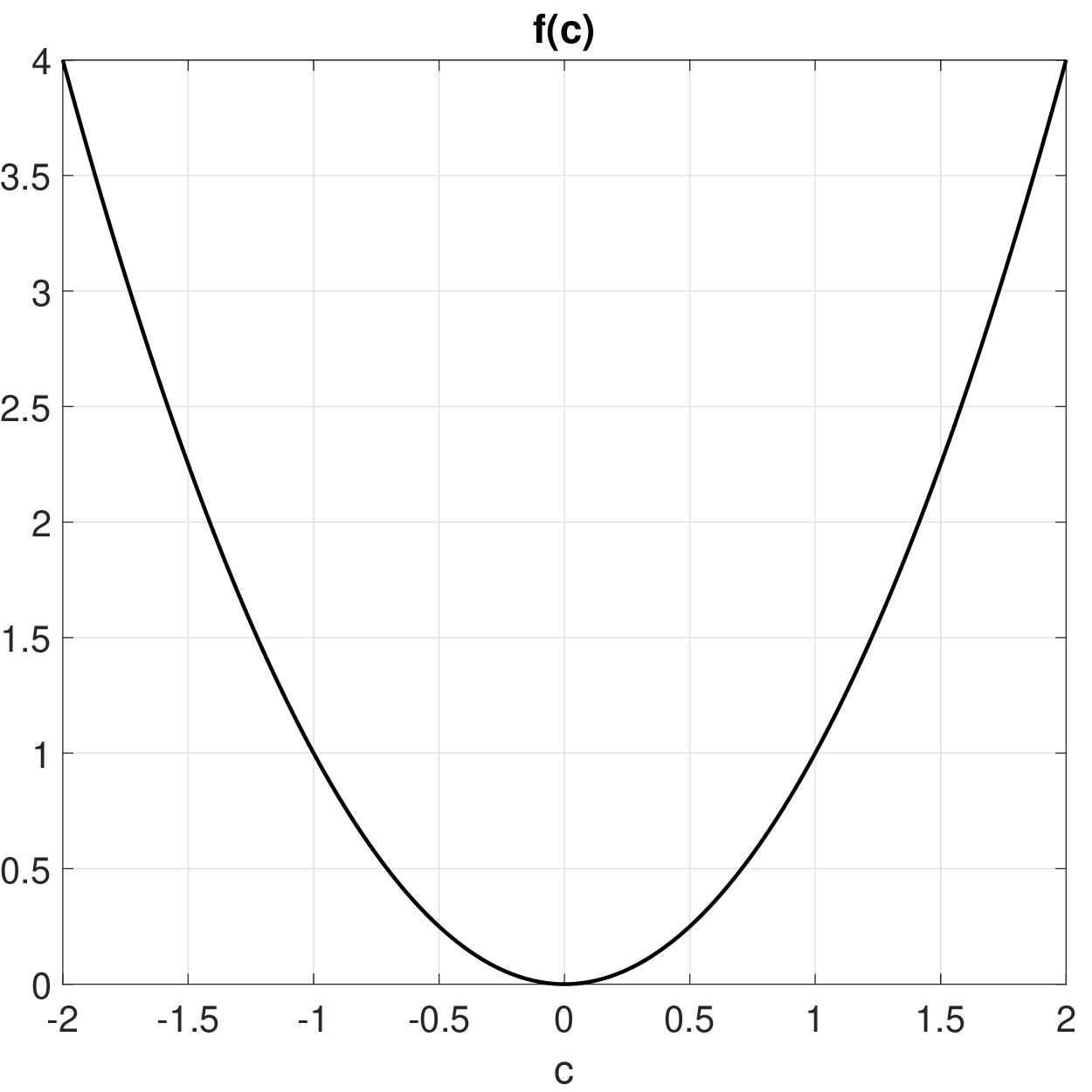}}
		\subfigure[]{
			\includegraphics[width=0.45\columnwidth]{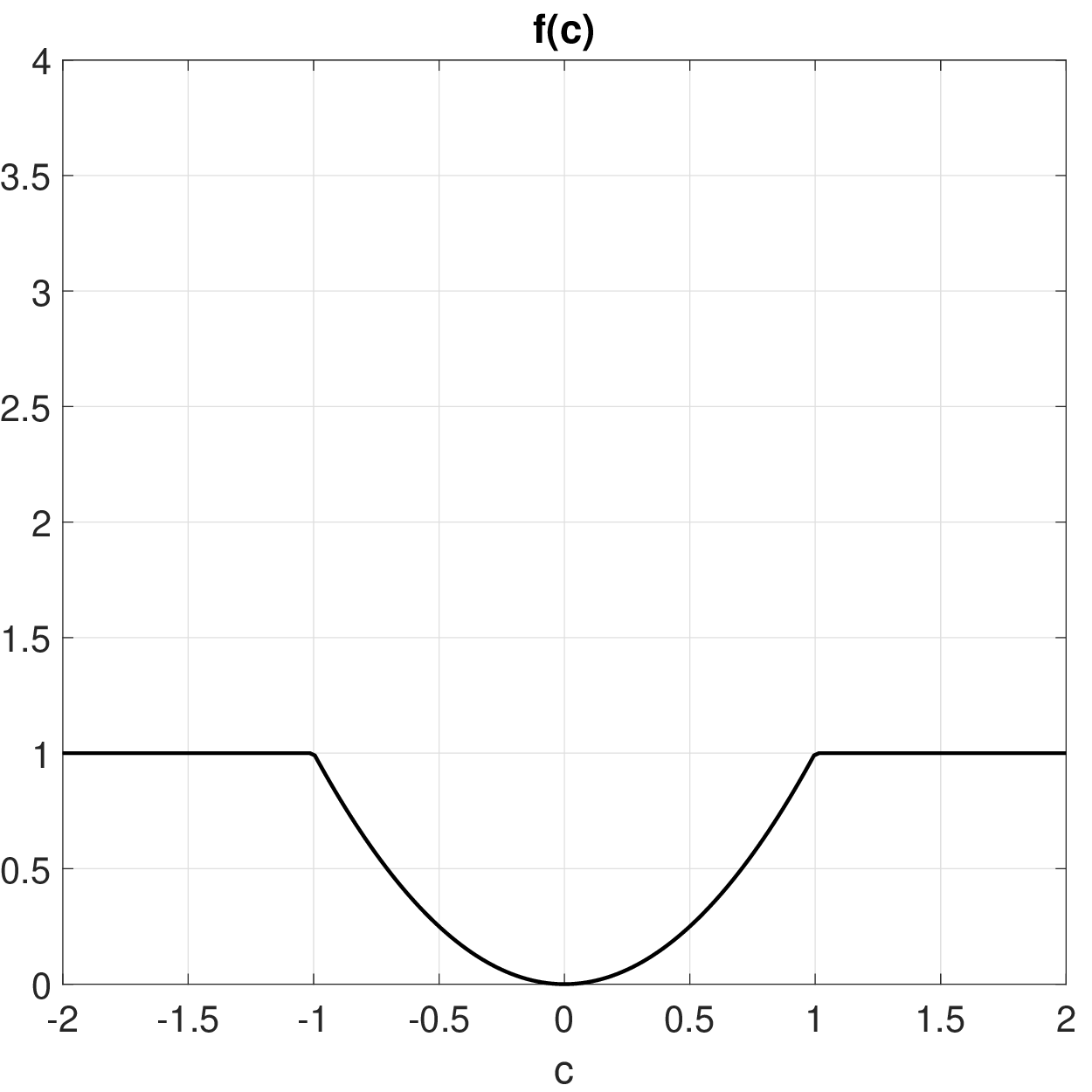}}
		\caption{{   The cost function for regular least squares (\textbf{a}) and for truncated least squares (\textbf{b}) with $\theta=1$. The~latter significantly reduces the associated cost of outliers, thus removing their contribution to the actual cost function to be optimized.}}
		\label{fig:robust.ls}
	\end{figure}

	Furthermore, this robust least-squares formulation can be shown to be exactly equivalent to an maximum a posteriori (MAP) probabilistic estimator if observations are assumed to be corrupted with additive Gaussian noise. 
	To prove this, we start from the formulation of a MAP estimator:
	\begin{eqnarray}
	\label{eq:map.estimator}
	\boldsymbol{x}^\star &=& \arg\max_\mathbf{x} p(z|\boldsymbol{x}) = \arg\max_\mathbf{x} \log (p(z|\boldsymbol{x})),
	\end{eqnarray}
	
	\noindent where, for~simplicity of notation, we used $z=\{z_1,\dots, z_N\}$ and $\boldsymbol{x}$
	to refer to the set of $N$ observed points and the vehicle pose for an arbitrary time step of interest, and~we took logarithms (a monotonic function that does not change the found optimal value) for convenience in further derivations.
	Assuming the following generative model for observations:
	\begin{eqnarray}
	\label{eq:generative.model}
	z_i &\sim& \bar{\mathbf{m}}^i + \mathcal{N}(\mathbf{0},\Sigma^i), \\
	\bar{\mathbf{m}}^i &=& \mathbf{p}_{map}^{i} \ominus \mathbf{x},  \\
	\Sigma^i &=& diag(\sigma^2,\sigma^2,\sigma^2),
	\end{eqnarray}
	
	\noindent where $p \ominus x$ means the local coordinates of point $p$ 
	as seen from the frame of reference $\boldsymbol{x}$, 
	$\mathcal{N}(\mathbf{m},\Sigma)$ 
	is the multivariate Gaussian distribution with mean $\mathbf{m}$ and variance $\Sigma$,
	and $\sigma$ is the standard deviation of the assumed additive Gaussian error in measured points. 
	Then, replacing Equation~(\ref{eq:generative.model}) into 
	Equation~(\ref{eq:map.estimator}), using the 
	known exponential formula for the Gaussian distribution, we find:
	\begin{scriptsize}
	\begin{eqnarray}
	\label{eq:deriv.start}
	\mathbf{x}^\star 
	&=& \arg\max_\mathbf{x} \left\{ \log (p(z|\mathbf{x}))\right\} \\
	&=& \arg\max_\mathbf{x} \left\{ \log \left(\prod_{i=1}^N p(z_i|\mathbf{x})\right)\right\} \quad \text{(Statistical independence of $N$ observations)} \\
	&=& \arg\max_\mathbf{x} \sum_{i=1}^N \left\{ \log \left(p(z_i|\mathbf{x})\right)\right\}\\
	&=& \arg\max_\mathbf{x} \sum_{i=1}^N \log \left( \left((2 \pi)^3 |\Sigma|\right)^{-1/2} \exp\left( -\frac{1}{2} (z_i - \bar{\mathbf{m}}^i)^\top \Sigma (z_i - \bar{\mathbf{m}}^i) \right) \right)  
	\\
	&=& \arg\max_\mathbf{x} \sum_{i=1}^N 
	\underbrace{ \cancel{ \log \left( \left((2 \pi)^3 |\Sigma|\right)^{-1/2} \right)} }
	_{\text{Does not depend on $\mathbf{x}$}}
	+ \cancel{\log} \left( \cancel{\exp}\left( -\frac{1}{2} (z_i - \bar{\mathbf{m}}^i)^\top \Sigma (z_i - \bar{\mathbf{m}}^i) \right) \right) \\
	&=& \arg\max_\mathbf{x} \sum_{i=1}^N 
	-\frac{1}{2} (z_i - \bar{\mathbf{m}}^i)^\top \Sigma (z_i - \bar{\mathbf{m}}^i) \\
	&=& \arg\min_\mathbf{x} \sum_{i=1}^N 
	\underbrace{\cancel{\frac{1}{2}}}_{\text{Constant}} (z_i - \bar{\mathbf{m}}^i)^\top \Sigma (z_i - \bar{\mathbf{m}}^i) \\
	\label{eq:deriv.end}
	&=& \arg\min_\mathbf{x}  
	\sum_{i=1}^N \left|\left|\frac{z_i - \bar{\mathbf{m}}^i}{\sigma}\right|\right|^2
	\quad \text{(Since $\Sigma$ is diagonal).}
	\end{eqnarray}
	\end{scriptsize}
	
	Identifying the last line above with Equation~(\ref{eq:ls}), it is 
	clear that the MAP statistical estimator is identical to 
	a least squares problem with error terms $c_i=\frac{z_i - \bar{\mathbf{m}}^i}{\sigma}$. 
	By using a truncated Gaussian in Equation~(\ref{eq:generative.model}),
	i.e., by~modeling outliers as having a uniform probability density, 
	one can also show that the corresponding MAP estimator becomes the 
	robust least-squares problem in Equation~(\ref{eq:robust.ls}).
	
	Therefore, the~proposed observation likelihood function in Algorithm \ref{alg1}
	enables an estimator to find the most likely pose of a vehicle while being 
	robust to outlier observations, for~example, from~dynamic obstacles.
}


\section{{  Justification of Decimation as an Approximation to the Likelihood~Function}}
\label{sect:justify.decimation}

\begin{figure}[H]
	\centering
	\includegraphics[width=0.62\columnwidth]{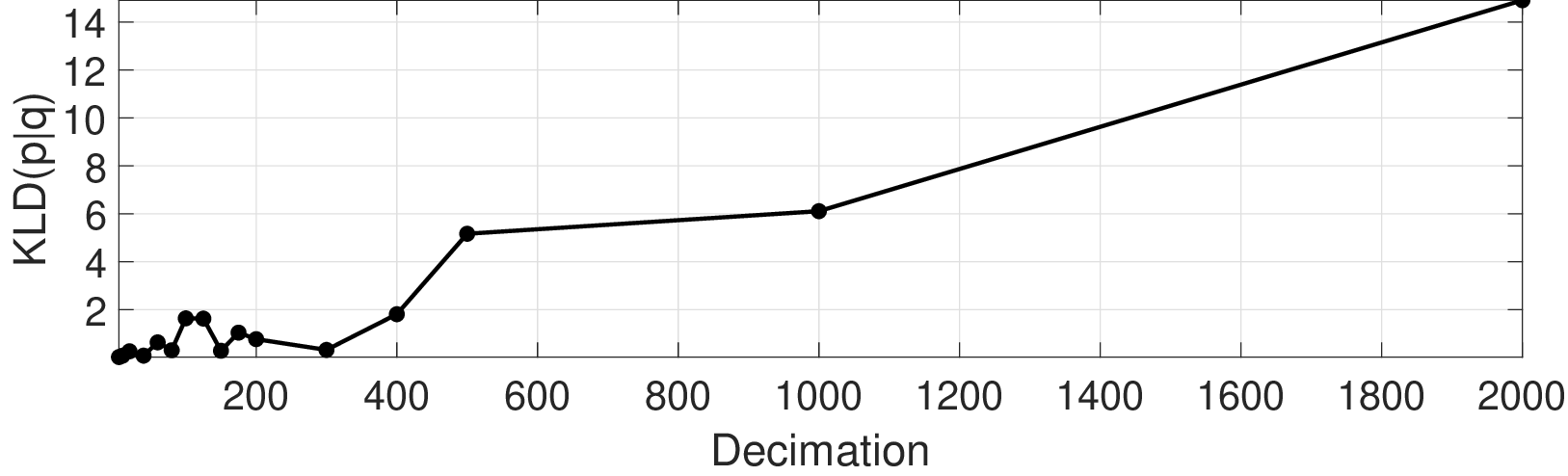}
	\caption{{Kullback--Leibler divergence (KLD) between the original and decimated version of the proposed likelihood model for point cloud observations. 
			Note that the decimated versions are remarkably similar to the original 
			one up to decimation ratios of roughly $\sim$500.
	}}
	\label{fig:kld}
\end{figure}

\begin{figure}[H]
	\centering
	\subfigure[]{
		\includegraphics[width=0.49\columnwidth]{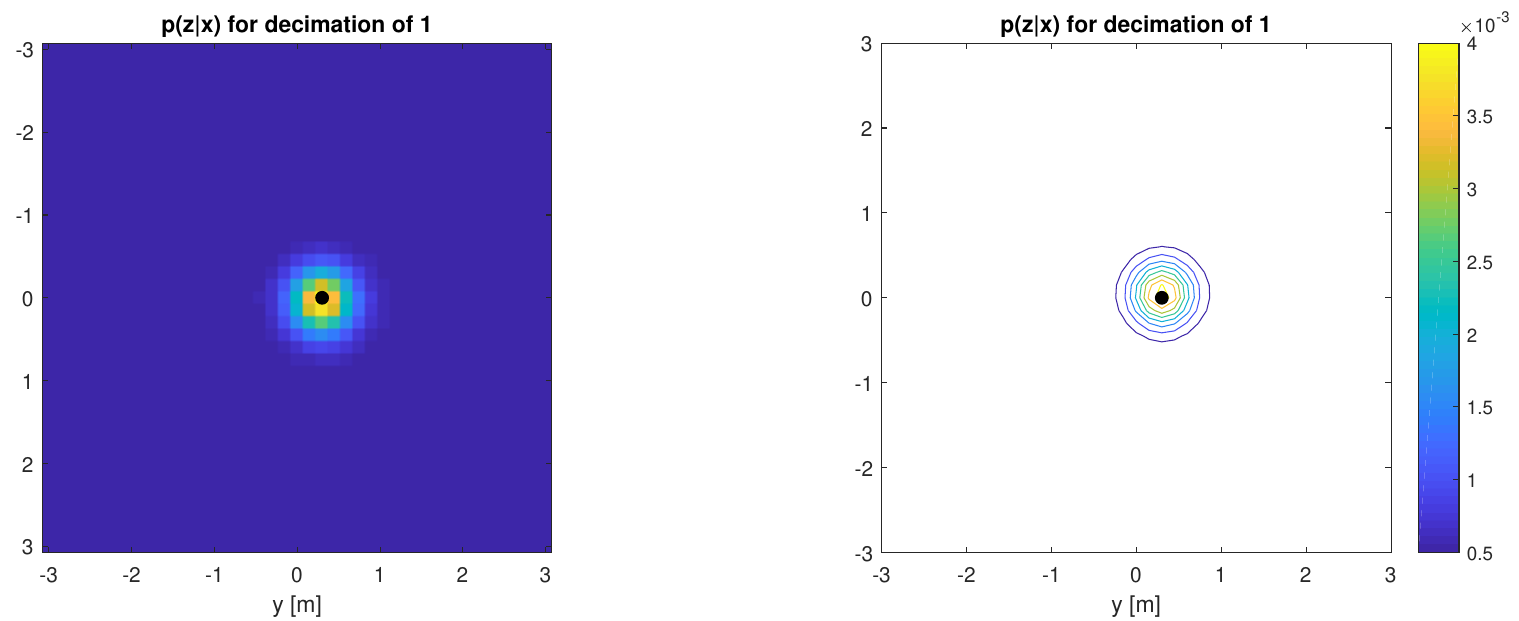}}
	\subfigure[]{
		\includegraphics[width=0.49\columnwidth]{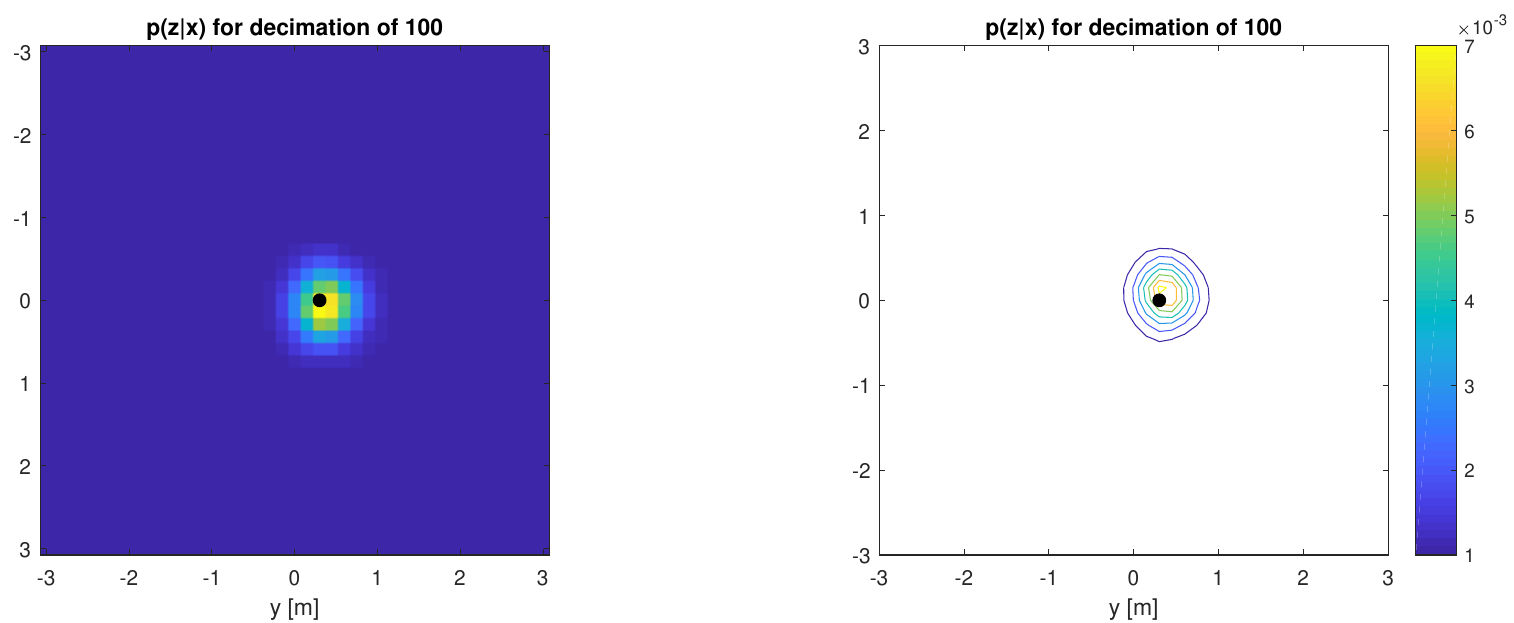}}
	\\
	\subfigure[]{
		\includegraphics[width=0.49\columnwidth]{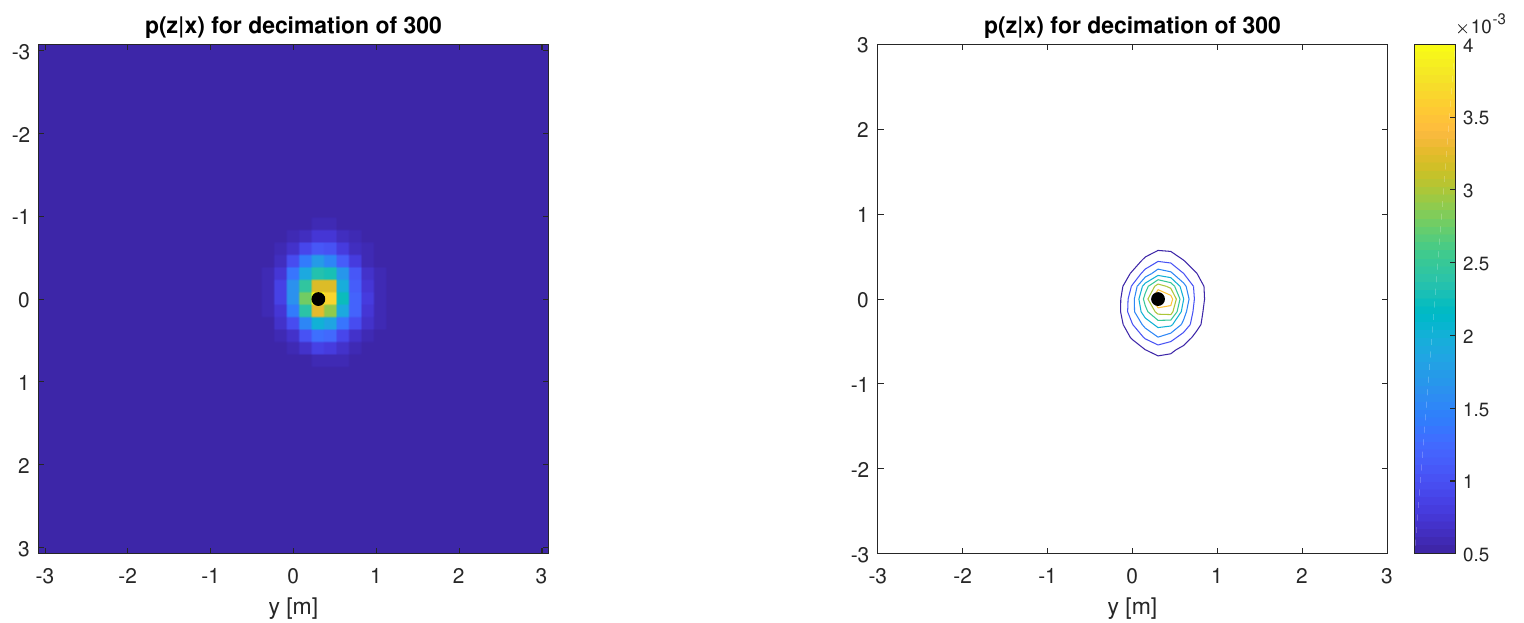}}
	\subfigure[]{
		\includegraphics[width=0.49\columnwidth]{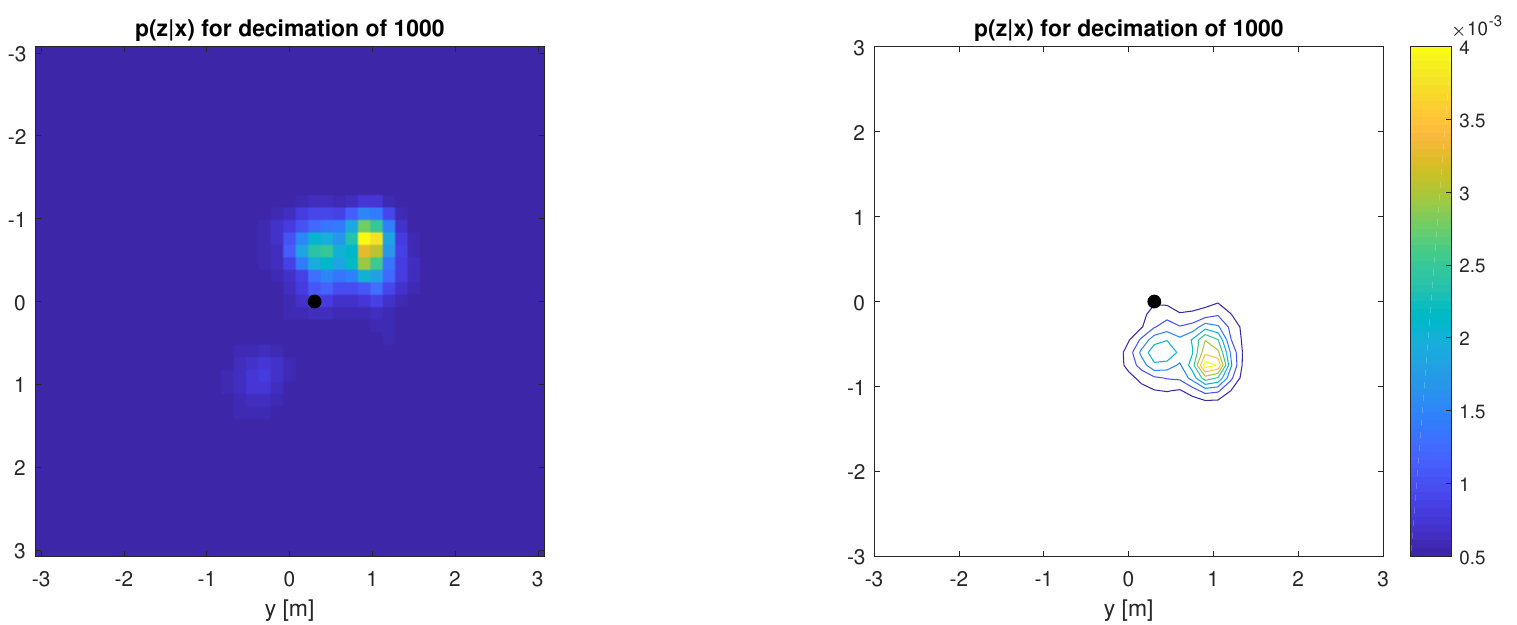}}
	\caption{{  Some planar slices of 6D likelihood functions evaluated for different decimation ratios, as~both an intensity color plot and a contour diagram. 
			Ground-truth vehicle pose is marked as a large block dot in all the figures. (\textbf{a}) Original p.d.f. $p(z|x)$; (\textbf{b}) Approximate $\hat{p}(z|x)$ for D = 100; (\textbf{c})~Approximate $\hat{p}(z|x)$ for D = 300; (\textbf{d}) Approximate $\hat{p}(z|x)$ for D = 1000.
	}}
	\label{fig:pdfs}
\end{figure}

{  
	A key feature of the proposed likelihood model in Algorithm \ref{alg1}, and~which is being 
	benchmarked in this work, is the decimation ratio, that is, how many points 
	from each observed scan are actually considered, with~the rest being plainly~ignored.
	
	The intuition behind this simple approach is that information in point clouds is highly redundant, such that, by~using only a fraction of the points, one could save a significant computational cost while still achieving good vehicle localization.
	From the statistical point of view, justifying the decimation is only possible 
	if the resulting likelihood functions (which in turn are probability density functions, p.d.f.) are still similar. 
	From Equations~(\ref{eq:deriv.start})--(\ref{eq:deriv.end}) above, solving the decimated problem is finding optimal pose $\hat{\boldsymbol{x}}^\star$ to the approximated p.d.f. with decimation ratio $D$:
	\begin{equation}
	\hat{\boldsymbol{x}}^\star 
	= 
	\arg\min_\mathbf{x}  
	\sum_{i=1,D,2D,\dots} \left|\left| \frac{z_i - \bar{\mathbf{m}}^i}{\sigma}\right| \right|^2.
	\end{equation}
	
	Numerically, the~decimated and the original p.d.f. are clearly not identical, but~
	this is not an issue for we are mostly interested in the location of the global optimum and the shape of the cost function in its neighborhood.
	The sum of convex functions is convex. In~our case, we have truncated 
	square cost functions (recall Figure~\ref{fig:robust.ls}b),
	but the overall p.d.f. will still be convex near the true vehicle pose. 
	Note that, since associations between observed and map points are 
	determined based solely on pairwise nearness, in~practice, the~observation likelihood 
	is not convex when evaluated far from the real vehicle pose. 
	However, this fact can be exploited by the particular kind of estimator used 
	in this work (particle filters) to obtain multi-modal pose estimations, where localization hypotheses are spread among several candidate ``spots''---for example, during~
	global relocalization, as~will be shown~experimentally.

	As a motivational example, we propose measuring the similarity between the decimated $\hat{p}(z|x_i)$ and the original  p.d.f. $p(z|x_i)$ using the Kullback--Leibler divergence (KLD) \cite{mackay2003information} using the following experimental procedure.
	Given a portion of the reference point cloud map, and~a scan observation from a Velodyne VLP-16,
	we have numerically evaluated both the original and the decimated likelihood 
	function in the neighborhood of the known ground-truth solution for the vehicle pose. 
	In particular, we evaluated the functions in a 6D grid (since SE(3) poses have six degrees of freedom) within an area of $\pm 3~ \unit{m}$
	for translation in $(x,y,z)$, $\pm 7.5$$^{\circ}$ for yaw (azimuth), and~$\pm 3$$^{\circ}$ for pitch and roll, 
	with spatial and angular resolutions of $0.15~ \unit{m}$ and $5$$^{\circ}$, respectively.
	For such discretized model of likelihood functions, we applied the discrete version 
	of KLD that is:
	\begin{equation}
	KLD(p||\hat{p}) = -\sum_i  p(z|x_i) \log\frac{\hat{p}(z|x_i)}{p(z|x_i)},
	\end{equation}

	\noindent which has been summed for the $3.1\times 10^6$ grid cells around the ground truth pose, for~a set of decimation ratio values.
	The result KLD is shown in Figure~\ref{fig:kld}, 
	and some example planar slices of the corresponding likelihood functions are illustrated in Figure~\ref{fig:pdfs}. 
	Decimated versions are clearly quite similar to the original one up to decimation ratios of roughly $\sim$500, which closely coincides with statistical localization errors presented later on in Section~\ref{sect:results}.
}
\unskip
\unskip
\unskip

\section{Experimental~Platform}
\label{sect:experimental}

\begin{figure}[H]
	\centering
	\includegraphics[width=0.83\columnwidth]{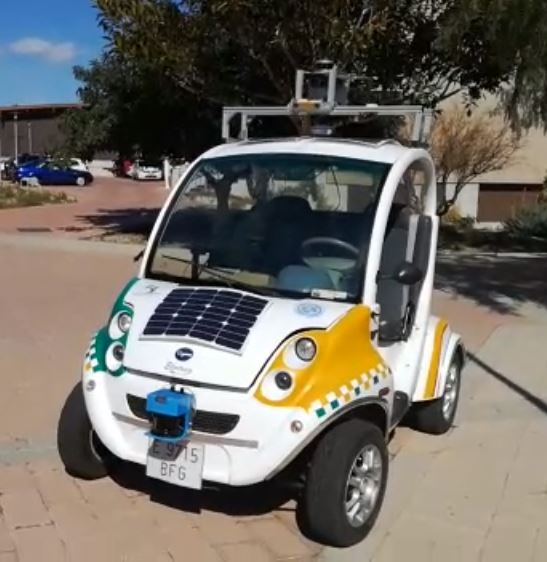}
	\caption{The autonomous vehicle prototype employed in this~work.}
	\label{fig:ecar.photo}
\end{figure}

{In order to perform the experimental test of this work, a~customized urban electric vehicle has been used (Figure \ref{fig:ecar.photo}).
	Among the sensors and actuators that have been installed in the vehicle are a} steering-by-wire system, which comprises a DC motor (Maxon RE50, Maxon Motor AG, Sachseln, Switzerland, diameter 50 mm, graphite brushes and 200 Watts) coupled to the conventional steering mechanism, commanded by a Pulse width modulation (PWM) 
signal and two encoders to close the control loop. The~main feedback sensor is an incremental encoder HEDL5540 with a resolution of 500 pulses per revolution, and~a redundant angle measurement is performed by an absolute encoder EMS22A with 10 bits of resolution. Another couple of encoders are mounted in the rear wheels to serve as odometry. Finally, the~prototype is equipped with a Novatel SPAN ING GNSS solution and a Velodyne VLP-16 3D LiDAR. {   More details about both mechanical characteristics and sensors placement are provided in~Appendix \ref{sect:veh.dataset}.}

The software architecture runs on top of a PC (64 bit Ubuntu GNU/Linux) and under Robotics Operative System (ROS) \cite{quigley2009ros} (version Kinetic),
with additional applications from the Mobile Robot Programming Toolkit \href{https://www.mrpt.org/}{(MRPT)} (\url{https://www.mrpt.org/}), version~1.9.9.

We provide open-source C++ implementations for all the modules employed in this work. Sensor~acquisition for Novatel GNSS and Velodyne scanners were implemented as C++ classes in the 
MRPT project. We also plan to release ROS wrappers in the repository \href{https://github.com/mrpt-ros-pkg/mrpt_sensors}{\texttt{mrpt\_sensors}} (\url{https://github.com/mrpt-ros-pkg/mrpt_sensors}). Our~implementation supports programmatically changing all Velodyne parameters (e.g., rpm), reading in dual range mode, etc. Odometry and low-level control are implemented in an \href{https://github.com/ual-arm-ros-pkg/ual-ecar-ros-pkg}{independent} (\url{https://github.com/ual-arm-ros-pkg/ual-ecar-ros-pkg}) ROS repository. Particle filter algorithms are also part of the MRPT libraries and have ROS wrappers in~\href{https://github.com/mrpt-ros-pkg/mrpt_navigation}{\texttt{mrpt\_navigation}} (\url{https://github.com/mrpt-ros-pkg/mrpt_navigation}).

\section{Results and~Discussion}
\label{sect:results}

Next, we discuss the results for each individual experiment and benchmark. All experiments ran within a single-thread on
an Intel i5-4590 CPU, (Intel Corporation, Santa Clara, CA, USA) @ 3.30~GHz. Due to the stochastic nature of PF algorithms, statistical results are presented for all benchmarks, which have been evaluated a number of times feeding the pseudorandom number generators with different~seeds.

\begin{figure}[H]
\centering
\subfigure[]{
	\includegraphics[width=0.73\columnwidth]{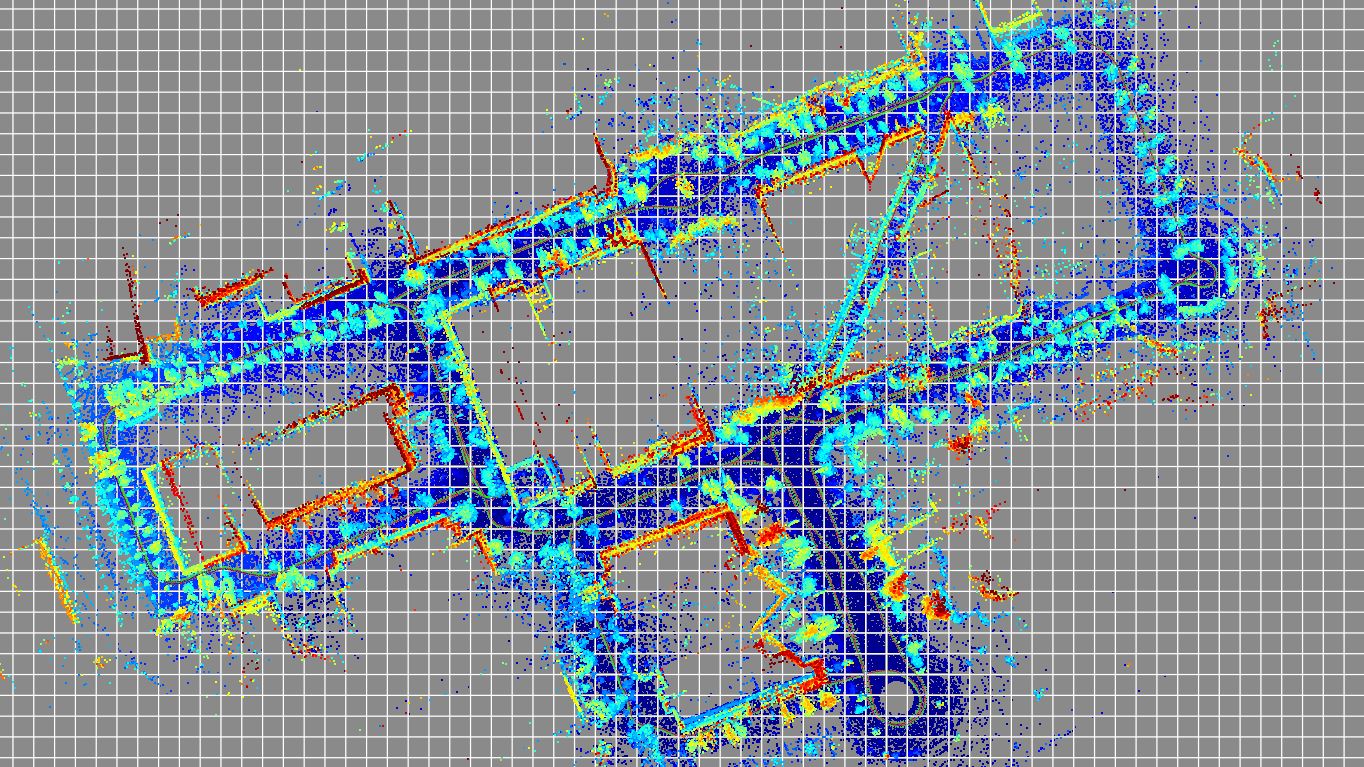}}
\subfigure[]{
	\includegraphics[width=0.73\columnwidth]{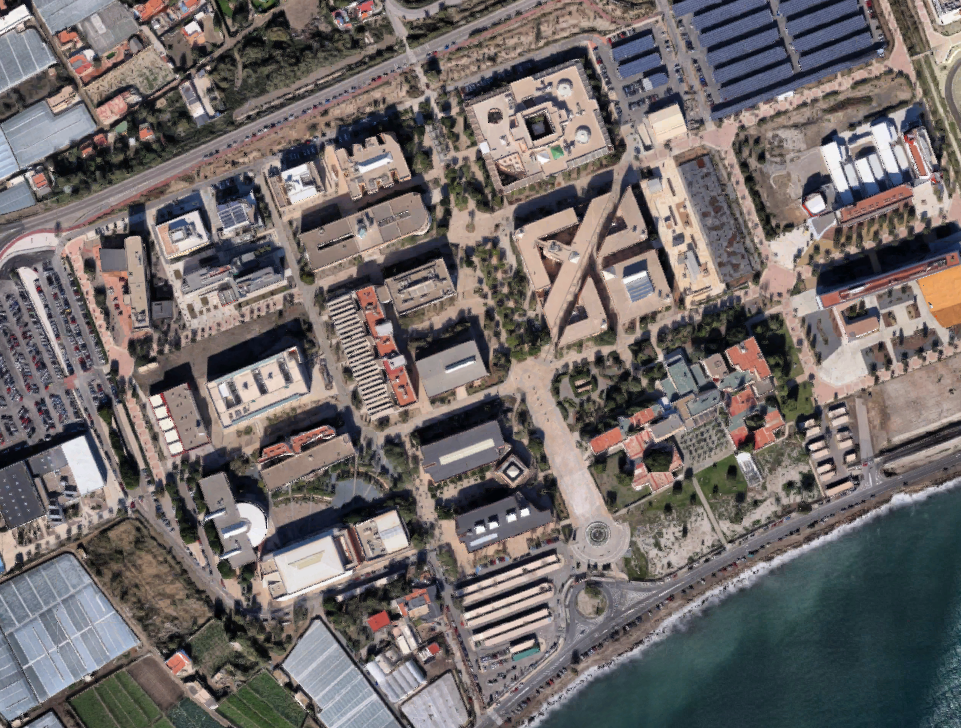}}
\caption{{Bird-eye view of (\textbf{a}) the ground-truth map used in the benchmark, along with (\textbf{b}) a corresponding satellite image of the UAL campus.}}\label{fig:results.map}
\end{figure}

\subsection{Mapping}
\begin{figure}[H]
	\centering
	\includegraphics[width=0.7\columnwidth]{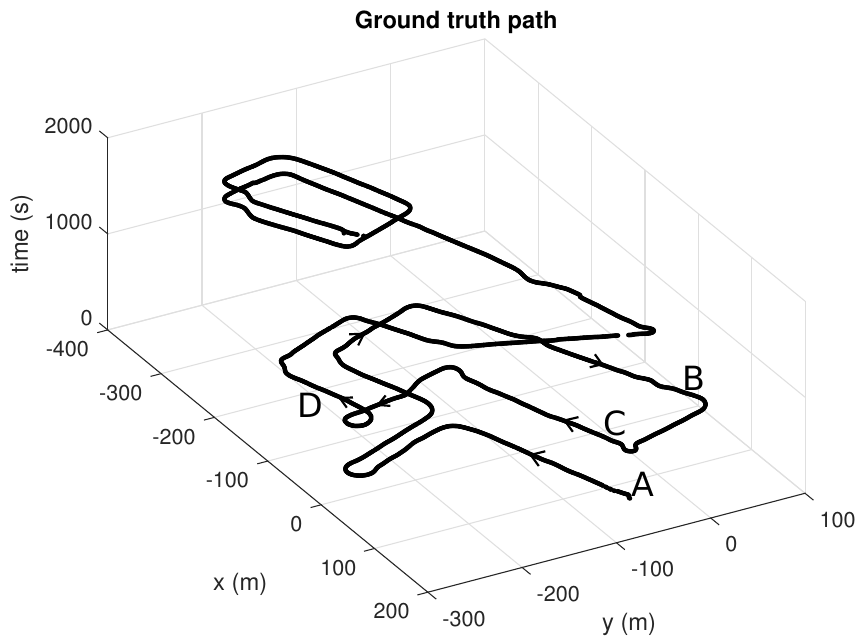}
	\caption{{   The whole ground-truth path of the presented dataset, with~time used as vertical axis to help identifying the loops. Please compare with Figure~\ref{fig:results.map} for reference.}}
	\label{fig:results.map.path}
\end{figure}

We acquired a dataset in the UAL campus (see Figure~\ref{fig:results.map}) with the purpose of serving to build a reference metric map and also to benchmark PF-based localization algorithms.
As described in former sections, we used centimeter-accurate GPS positioning and Novatel SPAN INS attitude estimation for orientation angles.
Poses were recorded at 20 Hz along the path shown in Figure~\ref{fig:results.map}b.
Since time has been represented in the vertical axis, it is easy to see how
the vehicle was driven through the same areas several times during the dataset.
In particular, we manually selected a first fragment of this dataset to generate a metric map (segment A--B in Figure~\ref{fig:results.map}b), then a second non-overlapping fragment (segment C--D) to test the localization algorithms as discussed in the following.
The global map obtained for the entire dataset is depicted in Figure~\ref{fig:results.map}a, whereas the corresponding ground-truth path can be seen in Figure~\ref{fig:results.map.path}.

\unskip

\subsection{Relocalization, {  Part 1: Aided by Poor-Signal~GPS}}
\label{sect:result.reloc}

Global localization, the~``awakening problem'' or ``relocalization'' are all
names of specific instances of the localization problem:
those in which uncertainty is orders of magnitude larger than during regular operation.
Depending on the case and available sensors, uncertainty may span a few square meters
within one room, or~an entire city-scale area.
Since our work addresses localization in outdoor environments, we will assume that
a consumer-grade, low-cost GPS device is available during the initialization
of the localization system.
To benchmark such a situation, we initialize the PF with
different number of particles (ranging from 20 to 4000)
spread over an area of $30 \times 30 $ $\text{m}^2$ that includes
the actual vehicle pose.
No clue is given for orientation (despite the fact that it might be easy to obtain from low-cost magnetic sensors)
and no GPS measurements are used in subsequent steps of the PF, whose only inputs are Velodyne scans and odometry readings.
{   The size of this area has been chosen to cover a typical worst-case GPS positioning data with poor precision, 
	that is, with~a large dilution of precision (DOP). Such a situation is typically found in areas where direct sight of satellites is blocked by obstacles (e.g., trees, buildings). 
	Refer to~\cite{dutt2009investigation} for an experimental measurement of such GPS positioning errors.}

Notice that the particle density is small even for the largest case (N = 4000, density is \mbox{$4000/900 \approx 4.4$ particles/m${}^2$}),
but the choice of the optimal-sampling PF algorithm makes it possible to successfully converge to the correct vehicle pose within
a few~timesteps.

\begin{figure}[H]
	\centering
	\subfigure[]{ \includegraphics[width=0.47\columnwidth]{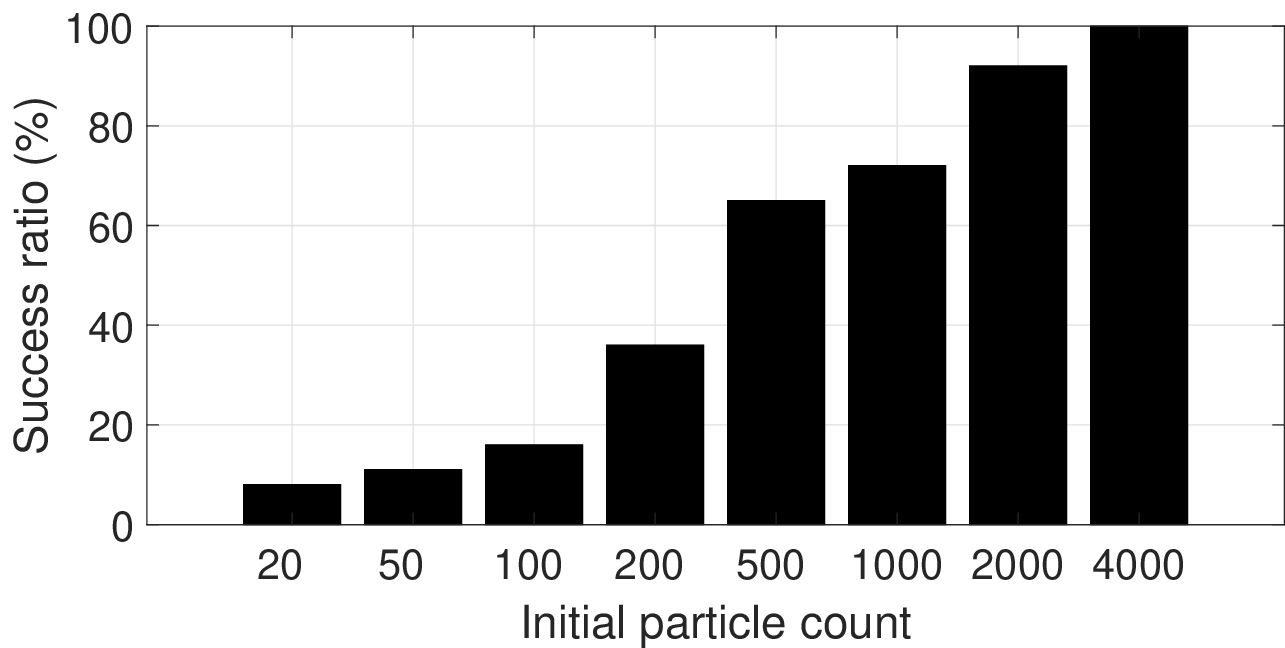}}
	\subfigure[]{ \includegraphics[width=0.47\columnwidth]{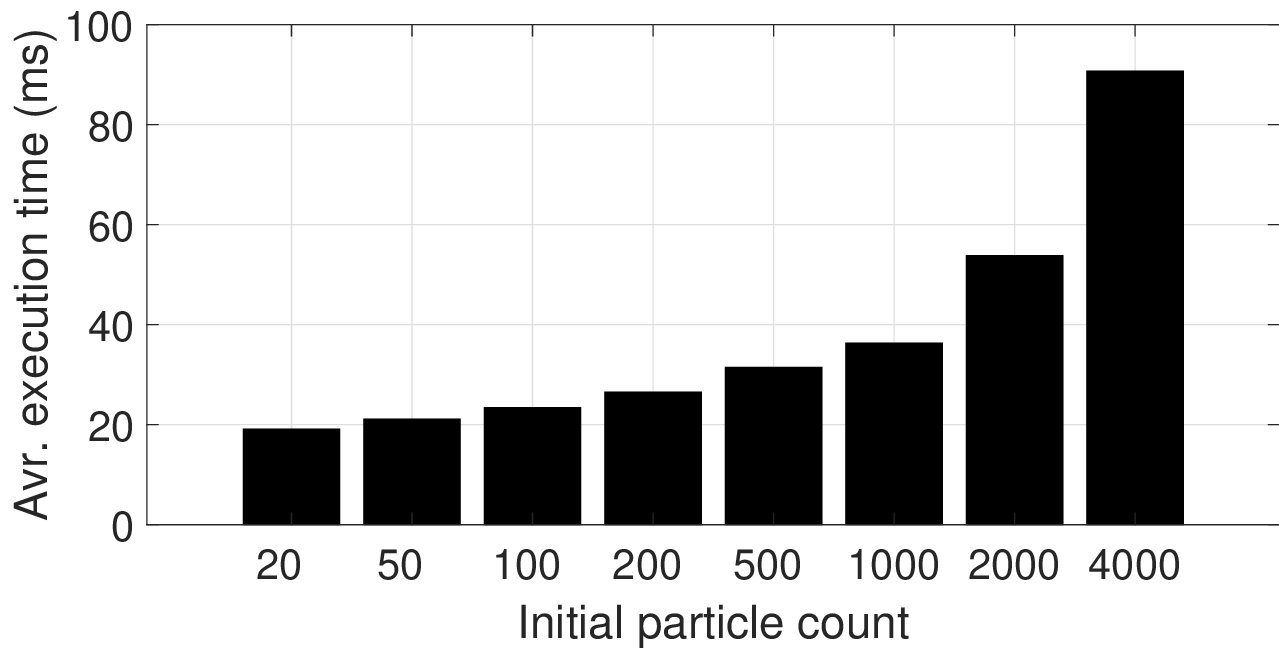}}
	\caption{Statistical results for the relocalization benchmark. Refer to Section
		\ref{sect:result.reloc} in the text for further details. (\textbf{a}) Success ratio of relocalization; (\textbf{b}) Computation cost.}
	\label{fig:results.global.success}
\end{figure}

We investigated what is the minimum particle density required to ensure a high probability of converging to
the correct pose, since oversizing might lead to excessive delays while the system waits for convergence.
Relocalization success was assessed by running a PF during 100 timesteps and checking whether
(i) the average particle pose is close to the actual (known) ground truth solution {   (closer than two meters)},
and (ii) the determinant of the covariance fitting all particles is below a threshold {   ($|\mathbf{\Sigma}|<2$)}.
Together, these conditions are a robust indicator of whether convergence was successful.
The experiment was run 100 times
for each initial population size $N$, 
{using a point cloud decimation of 100,}
and automatic sample size was in effect in the second and subsequent time steps.
The success ratio results can be seen in Figure~\ref{fig:results.global.success}a, and~demonstrate
that the optimal-sampling PF requires, in~our dataset, a~minimum of 4000 particles (4.4 particles/m${}^2$)
to ensure convergence. Obviously, the~computational cost grows with $N$ as
Figure~\ref{fig:results.global.success}b shows,
hence the interest in finding the minimum feasible population size.
Note that the computational cost is not linear with $N$ due to the complex evolution of the
actual population size during subsequent timesteps.
{  Normalized statistics regarding number of initial particles per area are also provided in Table~\ref{tab:results.density.convergence}, 
	where it becomes clear that an initial density of $\sim$2 particles/m$^2$ seems to be the minimum required to ensure convergence for the proposed model of observation likelihood.
}

\unskip

\subsection{{   Relocalization, Part 2: LiDAR~Only}}
\label{sect:result.reloc.global}

{   
	Next, we analyze the performance of the particle filter algorithm to localize, 
	from scratch, our~vehicle without any previous hint about its approximate pose 
	within the map of the entire~campus. 
	
	For that, we draw $N$ random particles following a uniform distribution (in $x$, $y$, and~also in the vehicle azimuth $\phi$) as the initial distribution, with~
	different values of $N$, and~after 100 time steps, we~detect whether the filter has converged to a single spot, and~whether the average estimated pose is actually close to the ground truth pose. The~experiment has been repeated 150 times for each initial particle count $N$. The~area where particles are initialized has a size of $420 \times 320\, \text{m}^2 = 134,400\,\text{m}^2$.
	Notice that the dynamic sample size algorithm ensures that computational cost quickly decreases as the filter converges, hence the higher computational cost associated with a larger number of particles only affects the first iterations (typically, less than 10 iterations). 
	\begin{table}[H]
		\centering
		\begin{small}
			\caption {{  Statistical results for the relocalization benchmark with an initial uncertainty area of $30 \times 30$~$\text{m}^2$. Refer to Section~\ref{sect:result.reloc}.}}
			\begin{tabular}{cc}
				\toprule
				\textbf{Initial Density (Particles/m$^2$)} & \textbf{Convergence Ratio}\\
				\midrule    
				0.02 & 21.7\% \\
				0.08 & 50.0\% \\
				0.17 & 68.3\% \\
				0.33 & 80.0\% \\
				0.83 & 96.7\% \\
				1.67 & 100\% \\
				3.33 & 100\% \\
				6.67 & 100\% \\
				\bottomrule
			\end{tabular}
			\label{tab:results.density.convergence}
		\end{small}
	\end{table}
	
	The summary of results can be found in Table~\ref{tab:results.global.success}, 
	and are consistent with the relationship between initial particle densities and convergence success ratio in Table~\ref{tab:results.density.convergence}.
	A video for a representative run of this test is available online for the convenience of the reader ({Video available in:} \url{https://www.youtube.com/watch?v=LJ5OV-KMQLA}).
}

\begin{table}[H]
	\centering
	\begin{small}
		\caption {{   Statistical results for the relocalization benchmark with an initial uncertainty area of the entire campus. Refer to Section~\ref{sect:result.reloc.global}}.}
		\begin{tabular}{ccc}
			\toprule
			\textbf{Initial Particle Count} & \textbf{Initial Density (Particles/m$^2$)} & \textbf{Convergence Ratio}\\
			\midrule
			1000 &	0.007	& 2.0\% \\
			2000 &	0.014	& 12.0\% \\
			5000 &	0.037	& 24.0\% \\
			10,000 &	0.074	& 32.0\% \\
			20,000 &	0.149	& 56.6\% \\
			30,000 &	0.223	& 63.3\% \\
			40,000 &	0.298	& 66.6\% \\
			50,000 &	0.373	& 75.3\% \\
			60,000 &	0.447	& 80\% \\
			70,000 &	0.522	& 81.3\% \\
			80,000 &	0.597	& 86.0\% \\
			100,000 &	0.746 &	89.3\% \\
			125,000 &	0.932 &	89.3\% \\
			150,000 &	1.119 &	92.6\% \\
			175,000 &	1.305 &	92.6\% \\
			200,000 &	1.492 &	96.0\% \\
			
			\bottomrule
		\end{tabular}
		\label{tab:results.global.success}
	\end{small}
\end{table}
\unskip

\subsection{Choice of PF~Algorithm}
\label{sect:results.pf.algs}

In this benchmark, we analyzed the pose tracking accuracy (positioning error with respect to ground-truth) and efficiency (average computational cost per timestep)
of a PF using the standard proposal distribution in contrast to another using the optimal proposal. 
{Please refer to~\cite{blanco2010ofn} for details on how this algorithm achieves a better random sampling of the target probability distribution, 
	by~simultaneously taking into account both the odometry model and the observation likelihood $p(z_t|x_t,m)$ in Algorithm \ref{alg1}.}

\begin{figure}[H]
	\centering
	\includegraphics[width=0.38\columnwidth]{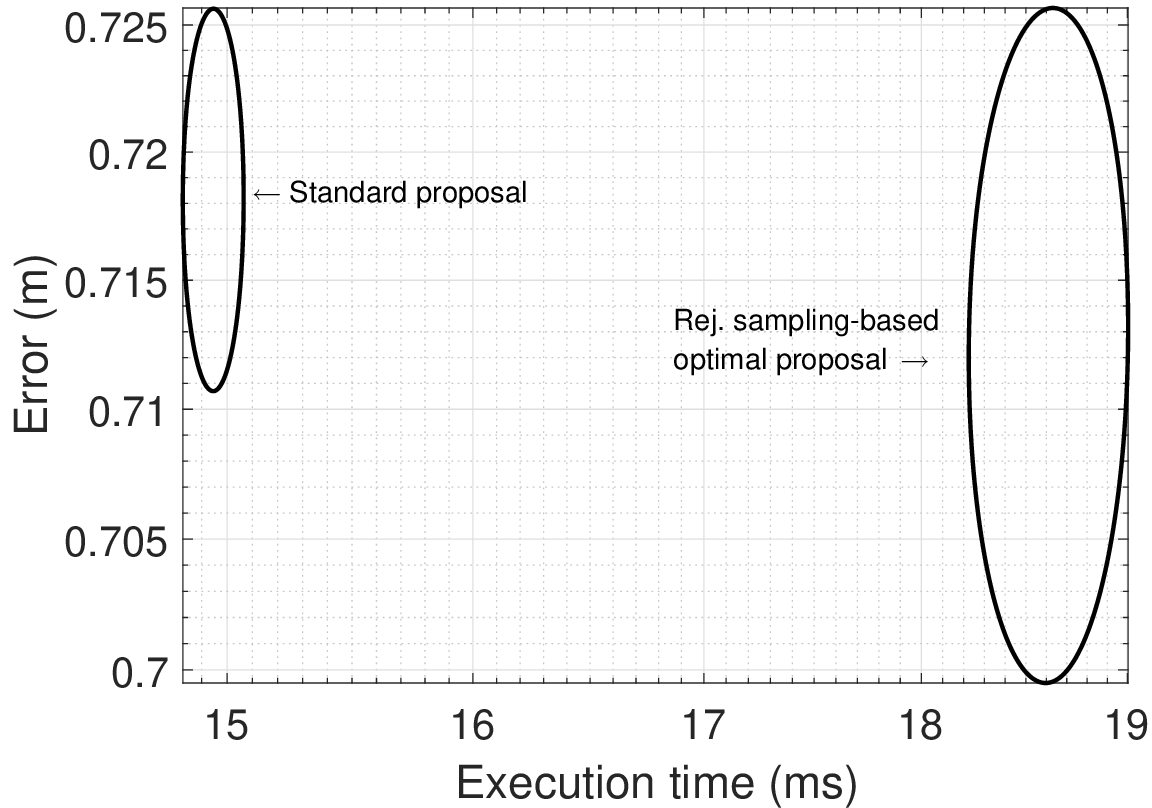}
	\caption{Execution time and average pose tracking error for two different PF
		algorithms. Ellipses~represent 95\% confidence interval as reconstructed from data of 25 repetitions of the same pose tracking experiment with different random seeds. Point cloud decimation was set to 100 in both algorithms. See~Section~\ref{sect:results.pf.algs} in the text for a discussion.}
	\label{fig:results.std.vs.opt}
\end{figure}

Experiments were run 25 times and average errors and execution times were collected for each algorithm,
then data fitted as a 2D Gaussian as represented in Figure~\ref{fig:results.std.vs.opt}.
The minimum population size of the standard PF was set to 200, while it was 10 for the optimal PF.
However, their ``effective'' number of particles are equivalent since each particle in the optimal algorithm
was set to employ 20~terations in the internal sampling-based stage.
The optimal PF achieves a slightly better accuracy with a relatively higher computational cost, which
still falls below 20 ms per iteration.
Therefore, the~conclusion is that the optimal algorithm is recommended, but~with a small practical gain,
a finding in accordance with previous works that revealed that the advantages of the optimal PF become
more patent when applied to SLAM, while only representing a substantial improvement for localization
when the sensor likelihood model is sharper~\cite{blanco2010ofn}.

\unskip

\subsection{Tracking~Performance}
\label{sect:results.tracking}

\begin{figure}[H]
	\centering
	\includegraphics[width=0.95\textwidth]{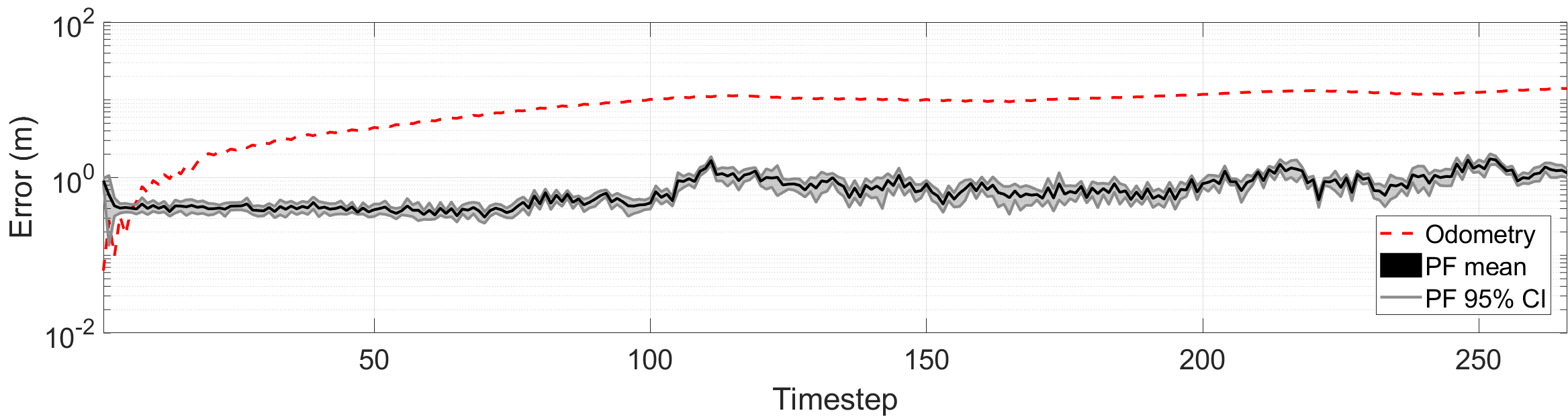}
	\caption{Pose tracking error and odometry-only error. See Section~\ref{sect:results.tracking} in the text for more~details.}
	\label{fig:results.tracking.error}
\end{figure}

To demonstrate the suitability of the proposed observation model,
we run 10 instances of a pose tracking PF using the standard proposal distribution,
point cloud decimation of 100, and~a dynamic number of samples with a minimum of 100.
We evaluated the mean and 95\% confidence intervals for the localization error over
the vehicle path, and~compared it to the error that would accumulate from odometry alone
in Figure~\ref{fig:results.tracking.error}.
As can be seen, the~PF keeps track of the actual vehicle pose with a
error median of 0.6 m (refer to Table~\ref{tab:Statistics_values}).

\unskip

\subsection{Decimating Likelihood~Evaluations}
\label{sect:results.decim}

\begin{figure}[H]
	\centering
	\includegraphics[width=1.0\textwidth]{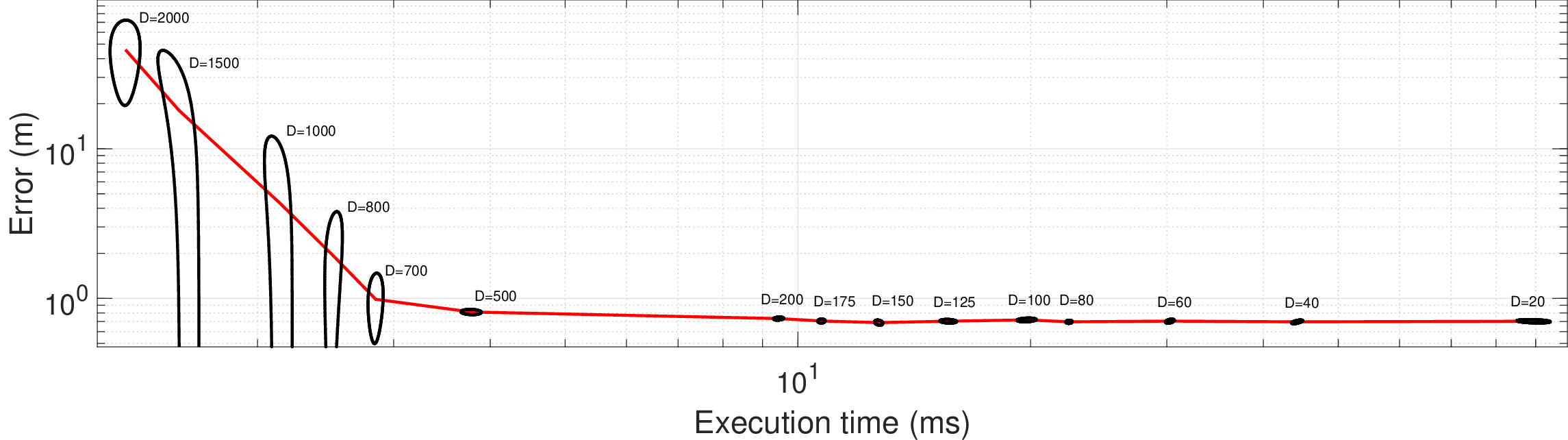}
	\caption{Positioning error and computational cost per timestep for different values of the likelihood function decimation parameter. Black: 95\% confidence intervals (ellipses deformed due to the logarithmic scale) for 25 experiments, red: mean values.
		Note that the initial error and uncertainty are larger than during the steady state of the tracking algorithm, since particles are initially uniformly-distributed over an area of $30 \times 30$ $\text{m}^2$.
		See Section~\ref{sect:results.decim} in the text for more details.}
	\label{fig:results.lik.decim}
\end{figure}

\begin{figure}[H]
	\centering
	\subfigure[]{ \includegraphics[width=0.28\columnwidth]{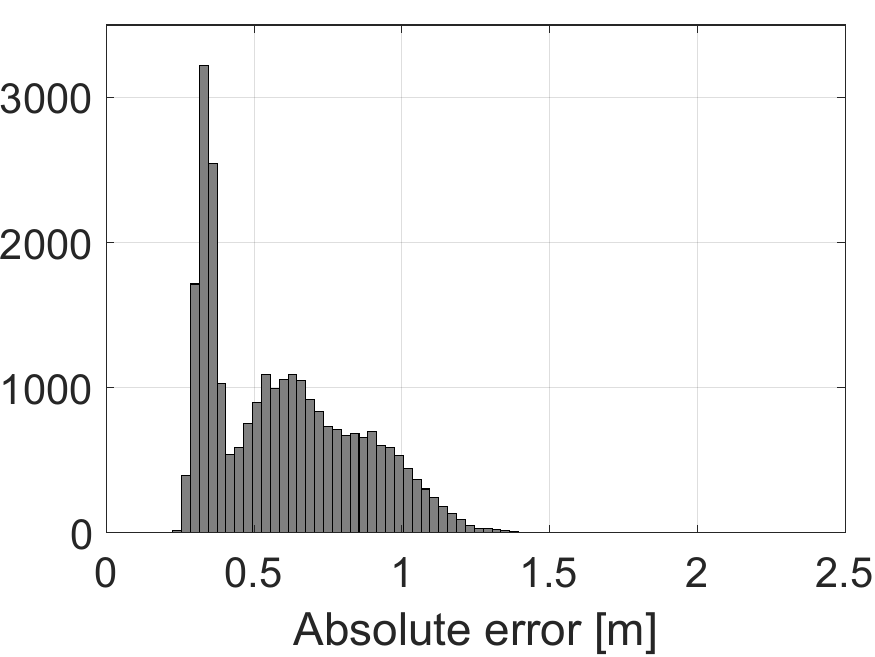}}
	\hspace{0.001\textwidth}
	\subfigure[]{ \includegraphics[width=0.28\columnwidth]{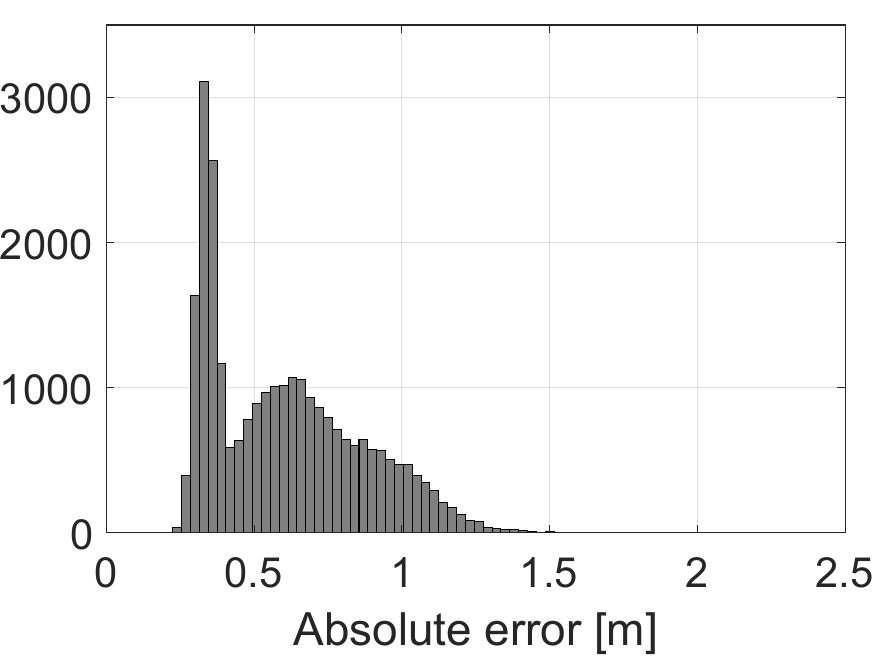}}
	\hspace{0.001\textwidth}
	\subfigure[]{ \includegraphics[width=0.28\columnwidth]{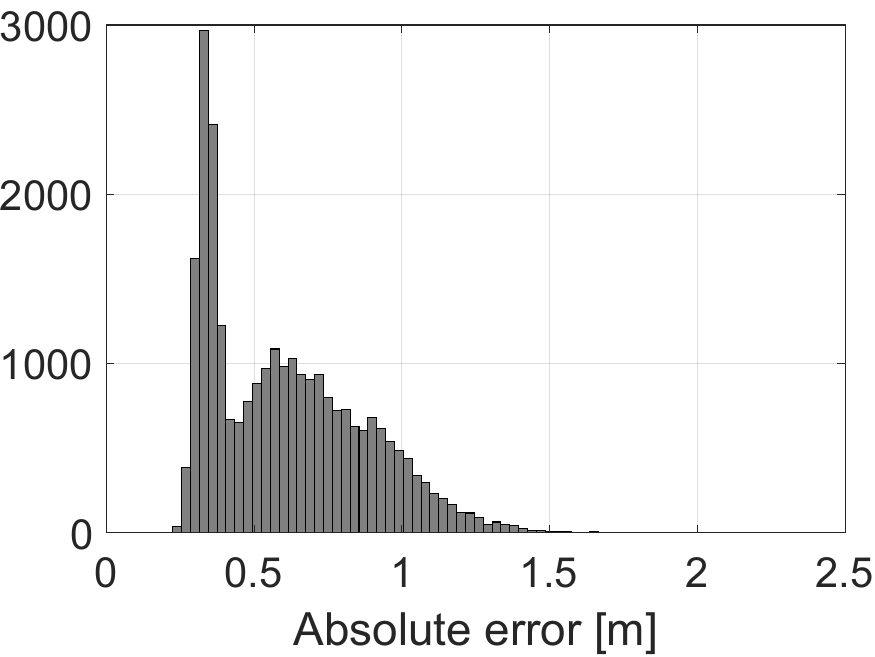}}
	\subfigure[]{ \includegraphics[width=0.28\columnwidth]{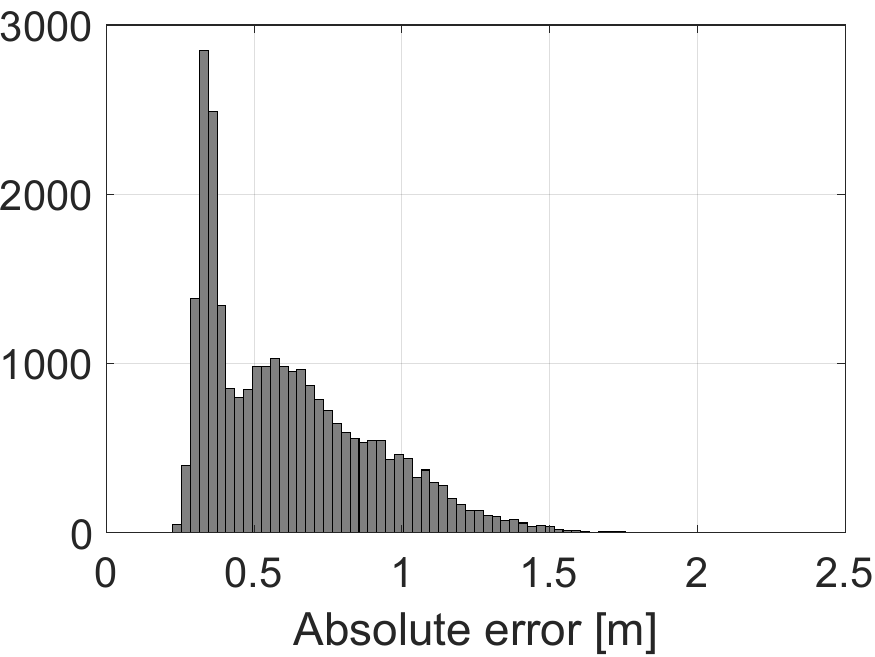}}
	\hspace{0.001\textwidth}
	\subfigure[]{ \includegraphics[width=0.28\columnwidth]{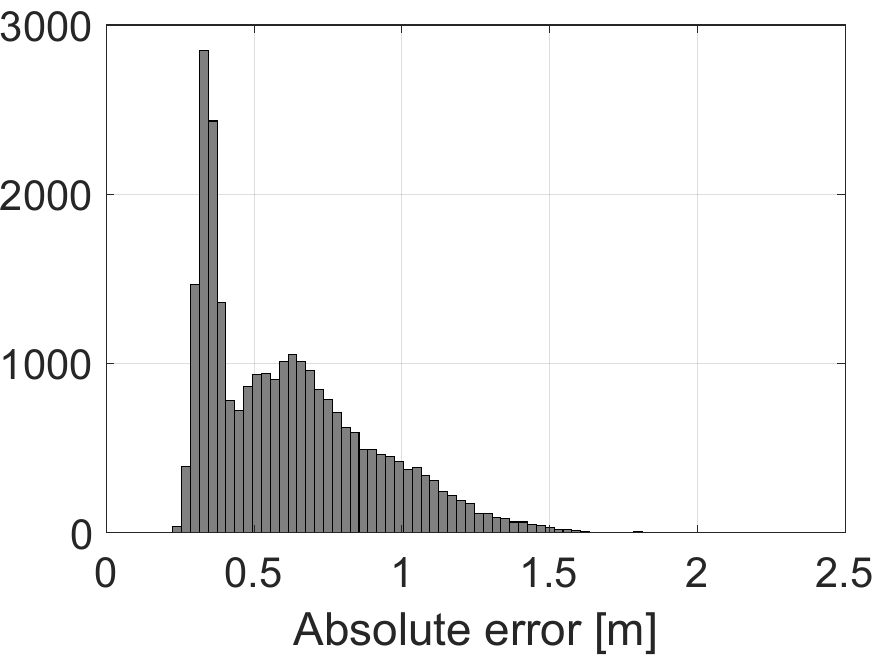}}
	\hspace{0.001\textwidth}
	\subfigure[]{ \includegraphics[width=0.28\columnwidth]{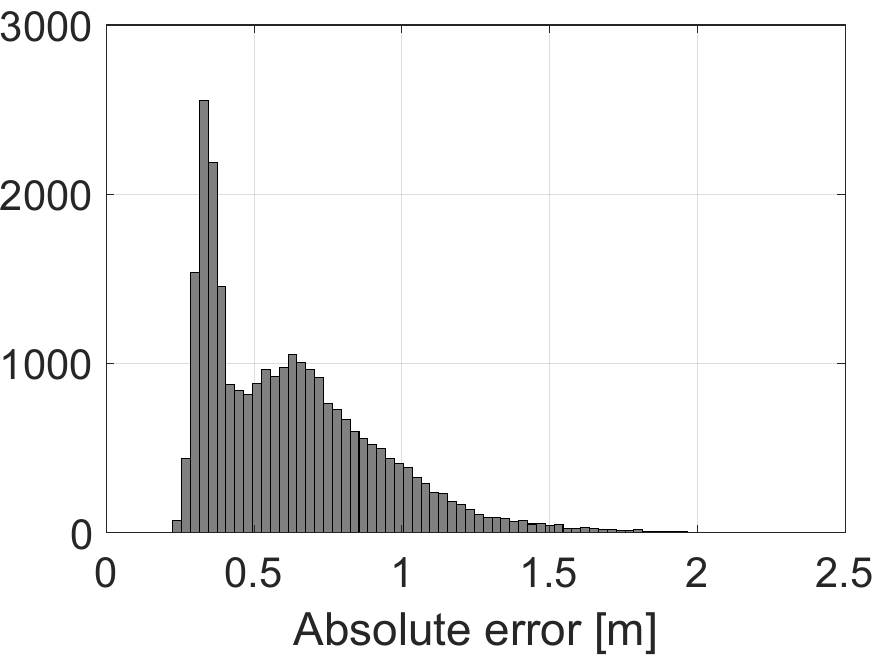}}
	\subfigure[]{ \includegraphics[width=0.28\columnwidth]{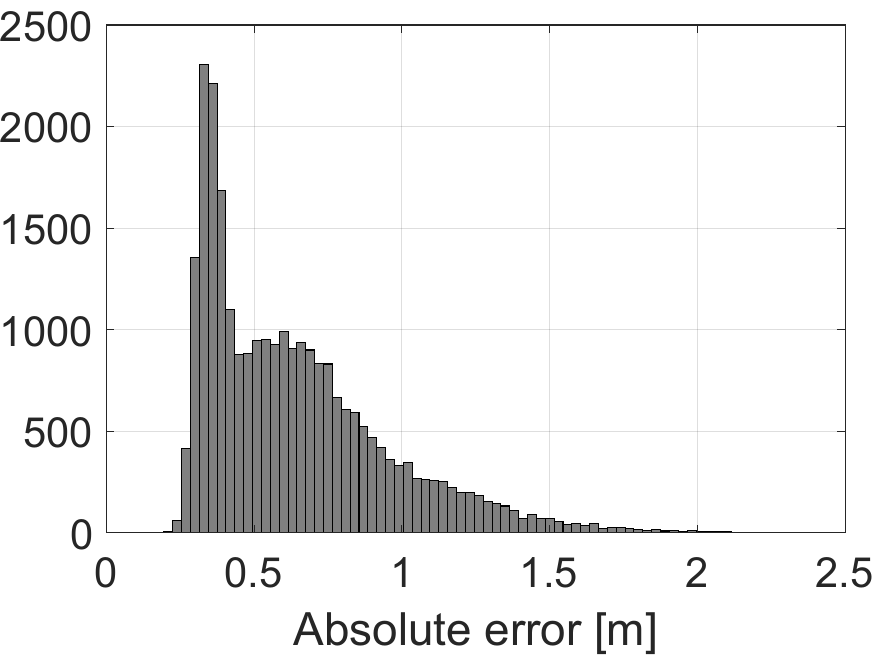}}
	\hspace{0.001\textwidth}
	\subfigure[]{ \includegraphics[width=0.28\columnwidth]{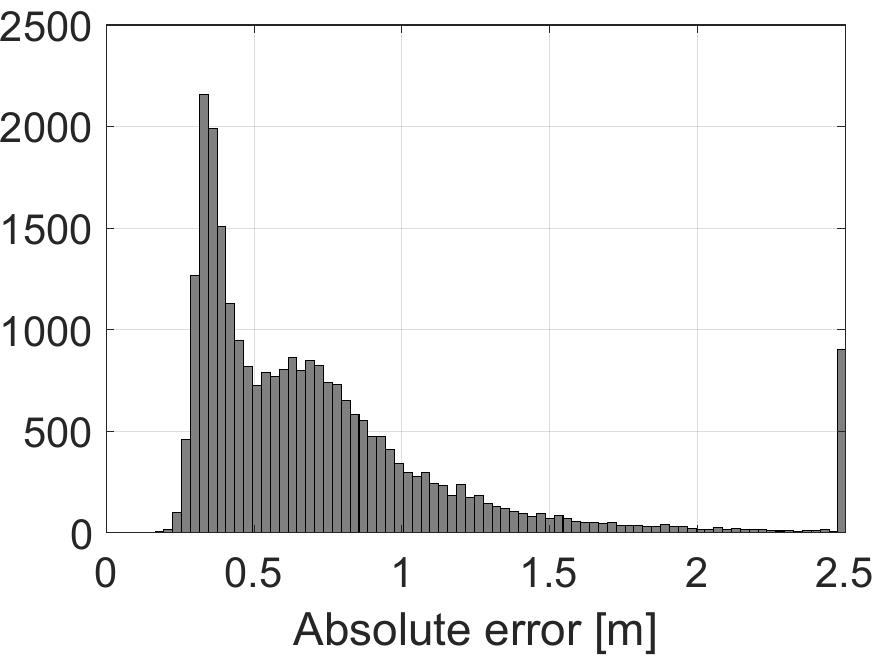}}
	\hspace{0.001\textwidth}
	\subfigure[]{ \includegraphics[width=0.28\columnwidth]{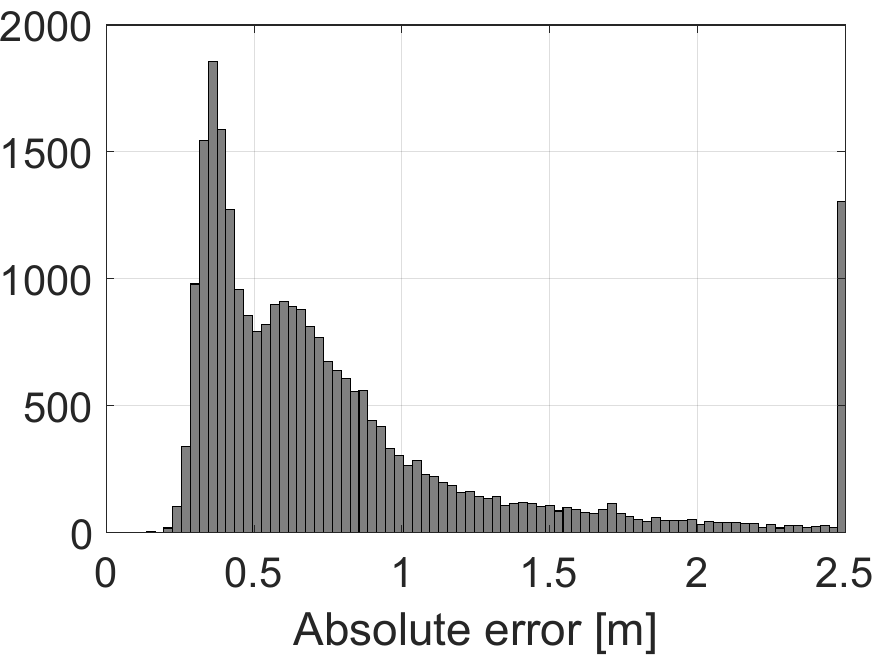}}
	\subfigure[]{ \includegraphics[width=0.28\columnwidth]{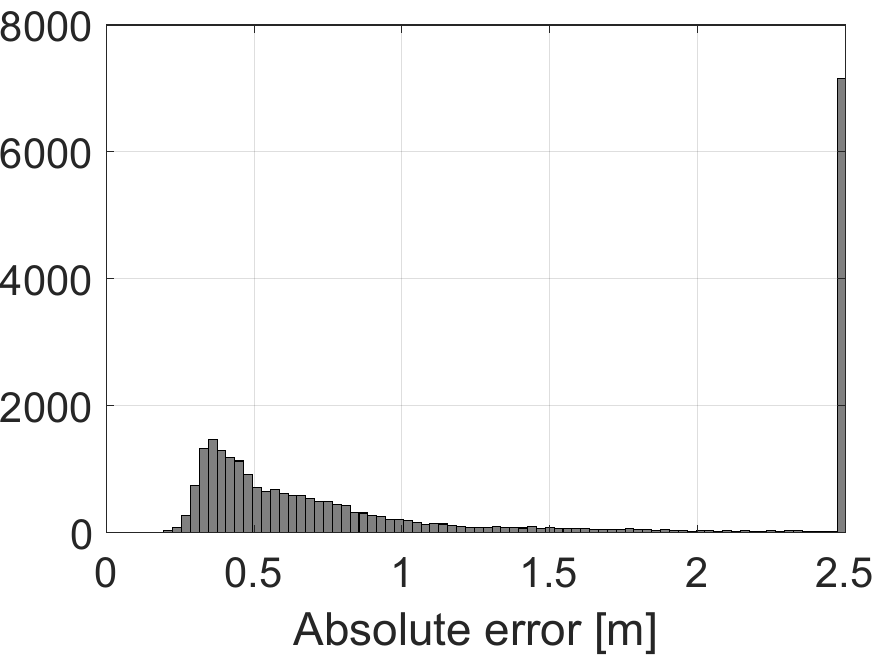}}
	\hspace{0.001\textwidth}
	\subfigure[]{ \includegraphics[width=0.28\columnwidth]{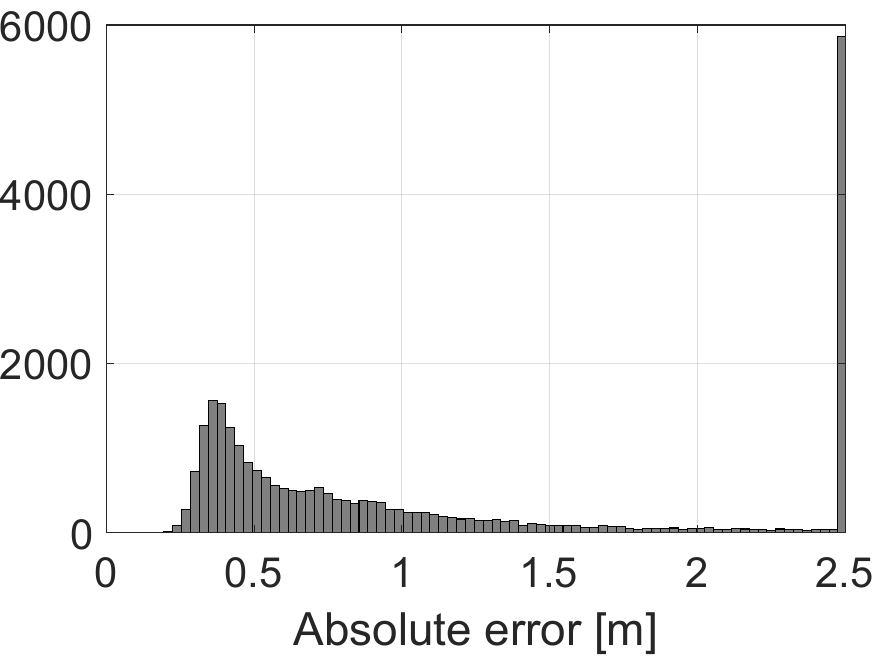}}
	\hspace{0.001\textwidth}
	\subfigure[]{ \includegraphics[width=0.28\columnwidth]{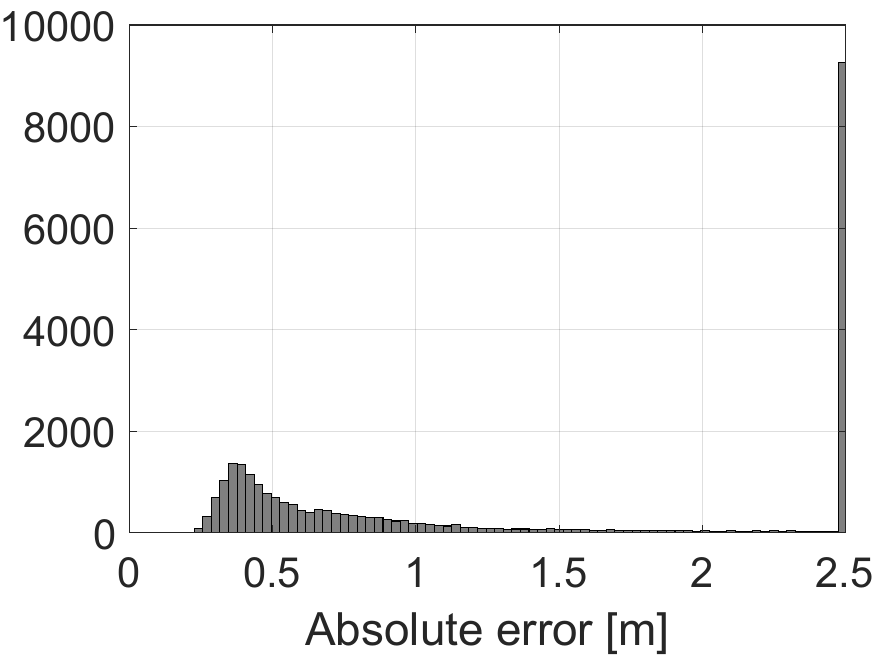}}
	\subfigure[]{ \includegraphics[width=0.28\columnwidth]{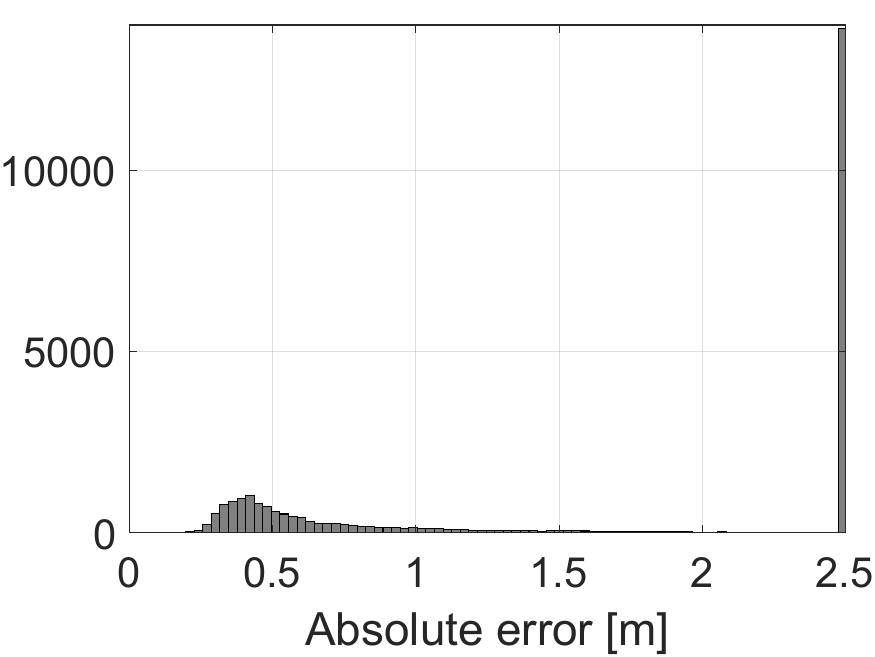}}
	\hspace{0.001\textwidth}
	\subfigure[]{ \includegraphics[width=0.28\columnwidth]{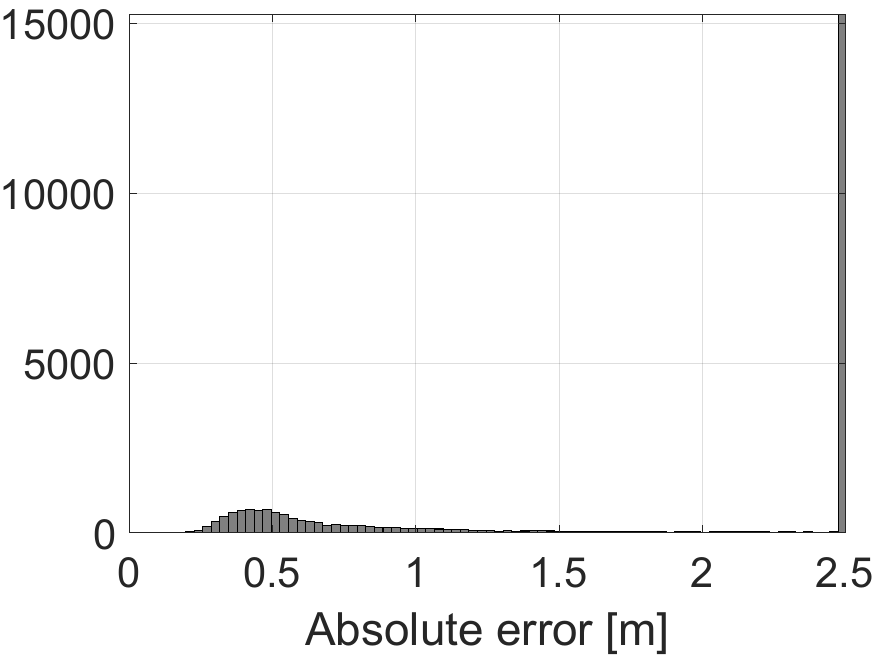}}
	\hspace{0.001\textwidth}
	\caption{Histograms of the pose tracking localization error for different decimation ratios. Note~that errors larger than 2.5 m are clipped into one single bin for the largest decimation ratios, for~the sake of providing a uniform horizontal scale in all~plots. (\textbf{a}) Decimation = 20; (\textbf{b}) Decimation = 40; (\textbf{c})~Decimation = 60; (\textbf{d}) Decimation = 80; (\textbf{e}) Decimation = 100; (\textbf{f}) Decimation = 150; (\textbf{g}) Decimation = 200; (\textbf{h}) Decimation = 300; (\textbf{i}) Decimation = 500; (\textbf{j}) Decimation = 700; (\textbf{k}) Decimation = 800; (\textbf{l})~Decimation = 1000; (\textbf{m}) Decimation = 1500; (\textbf{n}) Decimation = 2000}
	\label{fig:hist_decimations}
\end{figure}

Finally, we addressed the issue of how much information can be discarded
from each incoming scan while preventing the growth of positioning error.
Decimation is the single most crucial parameter regarding the computational cost
of pose tracking with PF, hence the importance of quantitatively evaluating
its range of optimal values. The~results, depicted in Figure~\ref{fig:results.lik.decim},
clearly show that decimation values in the range 100 to 200 should be the minimum
choice since error is virtually unaffected. In~other words, Velodyne scans apparently
have so much redundant information that we can keep only 0.5\% of them and still
remain well-localized. Statistical results of these experiments, and~the corresponding error histograms,
are shown in Table~\ref{tab:Statistics_values} and Figure~\ref{fig:hist_decimations}, respectively.
As can be seen from the results, the~average error is relatively stable for decimation values
of up to 500, and~quickly grows~afterwards.

\unskip

\begin{table}[H]
	\centering
	\begin{small}
		\caption {{  Localization error statistics for different decimation ratios applied to the input sensory data.}}
		\begin{tabular}{cccc}
			\toprule
			\textbf{Decim.} & \textbf{Mean (m)} & \textbf{Median (m)} & \textbf{Standard Deviation (m)}\\
			\midrule    
			20 & 0.615 & 0.585 & 0.250 \\
			40 & 0.620 & 0.585 & 0.257 \\
			60 & 0.626 & 0.590 & 0.261 \\
			80 & 0.635 & 0.579 & 0.282 \\
			100 & 0.635 & 0.587 & 0.279 \\
			150 & 0.641 & 0.591 & 0.290 \\
			200 & 0.656 & 0.586 & 0.315 \\
			300 & 1.207 & 0.623 & 3.528 \\
			500 & 0.951 & 0.641 & 1.182 \\
			700 & 8.356 & 0.746 & 20.023 \\
			800 & 6.497 & 0.765 & 17.832 \\
			1000 & 8.915 & 0.892 & 17.405 \\
			1500 & 21.674 & 4.325 & 30.847 \\
			2000 & 30.895 & 9.638 & 37.628 \\
			\bottomrule
		\end{tabular}
		\label{tab:Statistics_values}
	\end{small}
\end{table}
\unskip

\section{Conclusions}
\label{sect:conclusions}
In this work, we proposed an observation model for Velodyne scans, suitable for use within a PF, which has been successfully validated experimentally. Benchmarks showed that the optimal-PF algorithm is preferable in general due to its superior accuracy during pose tracking and its suitability to cope with the relocalization problem with an exiguous density of particles.
Furthermore, one of the most remarkable results is the finding that PFs are robust enough to  keep track of a vehicle pose while decimating the input point cloud from a Velodyne sensor by factors of two orders of magnitude. Such an insight, together with the use of a KD-tree for efficient querying the reference map, allows for running an entire localization update step within 10 to 20~ms.

\vspace{6pt}

\section*{Acknowledgments}
This work was partly funded by 
	the National R+D+i Plan Project 
	DPI2017-85007-R of the Spanish Ministry of Economy, Industry, and~Competitiveness and European Regional Development Fund (ERDF) funds.

\appendix
\section{{  Vehicle Description and Raw~Dataset}} 
\label{sect:veh.dataset}

\begin{table}[H]
	\centering
	\begin{small}
		\caption {Main characteristics of the prototype vehicle.}
		\begin{tabular}{ll}
			\toprule
			\textbf{Mechanic Characteristics} & \textbf{Value} \\
			\midrule  Lenght $\times$ Width $\times$ Height & $2680 \times 1525 \times 1780$ $\text{mm}^3$ \\
			Wheelbase & $1830$ mm \\
			Front/rear track width & $1285/1260$ mm \\
			Weight without/with batteries & $472/700$ kg \\
			\midrule
			\textbf{Electric Characteristics} &  \textbf{Value}\\
			\midrule	DC motor XQ $-$ 4.3 &  $4.3$ kW\\
			Batteries (gel technology) & $8 \times 6$ V $-210$ Ah  \\
			Autonomy &  $90$ km\\
			\bottomrule
		\end{tabular}
		\label{tab:ualCarCharacteristics}
	\end{small}
\end{table}
\unskip

\begin{figure}[H]
	\centering
	\setcounter{subfigure}{0}
	\subfigure[]{ \includegraphics[width=0.49\columnwidth]{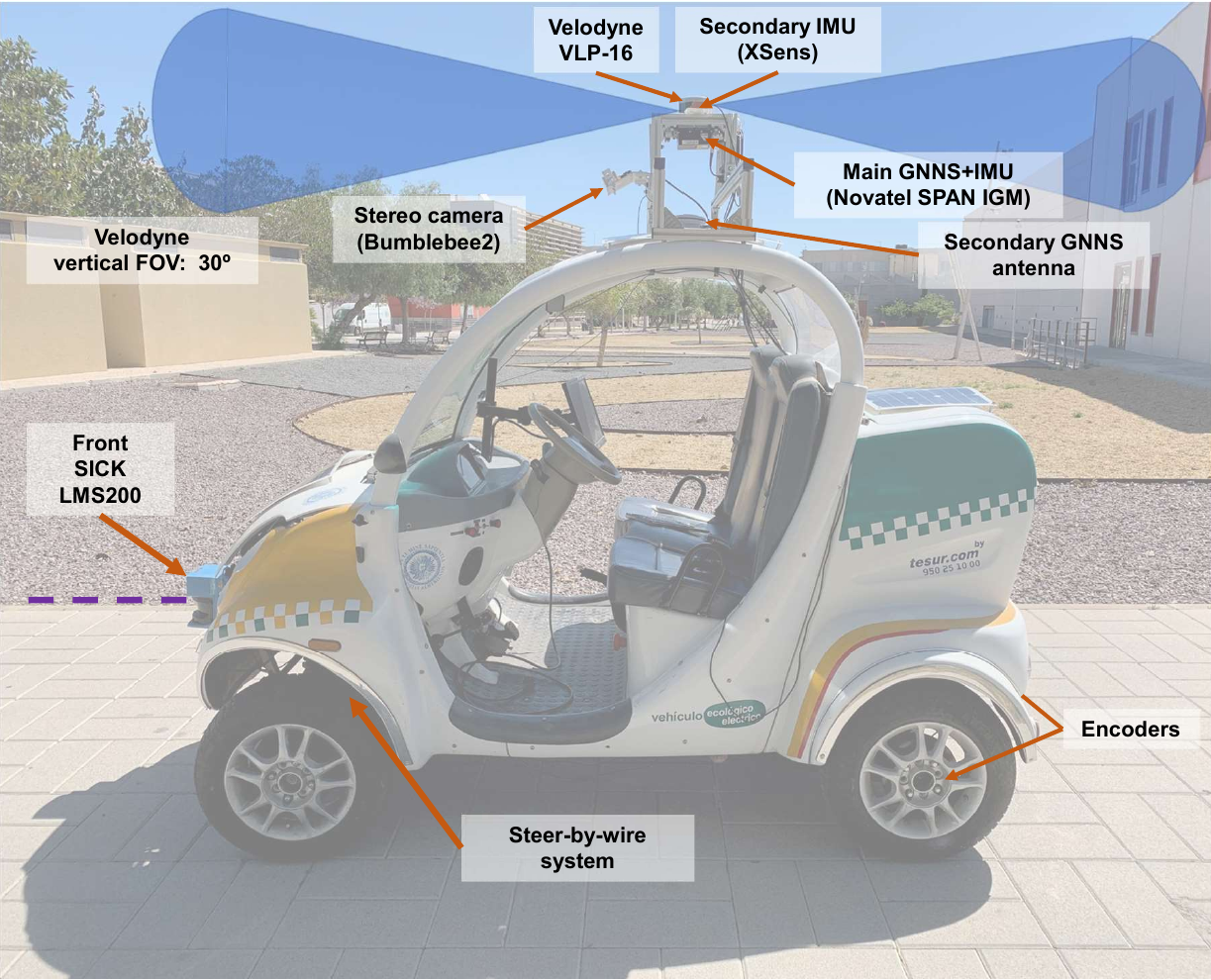}}
	\subfigure[]{ \includegraphics[width=0.49\columnwidth]{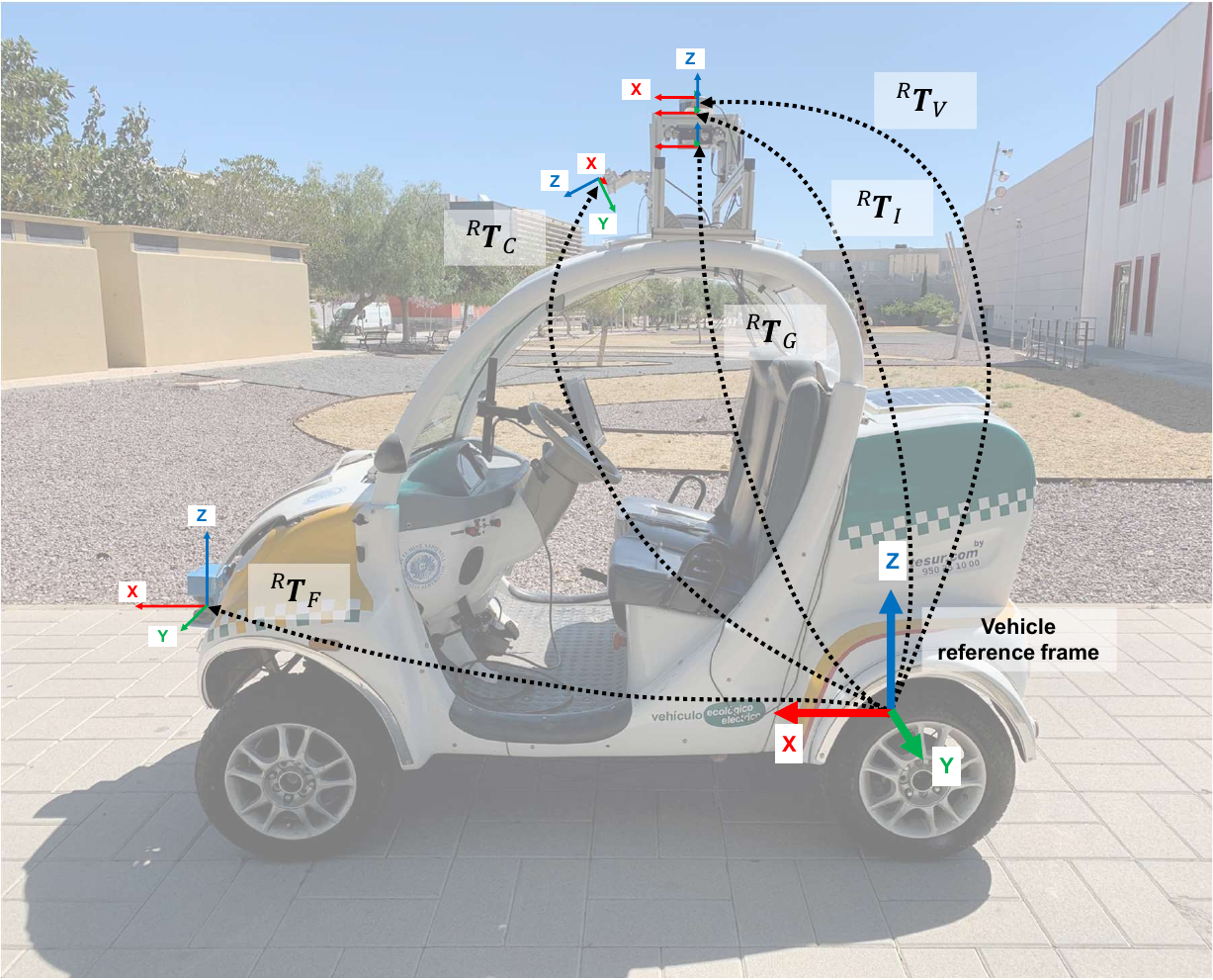}}
	\\
	\subfigure[]{ \includegraphics[width=0.49\columnwidth]{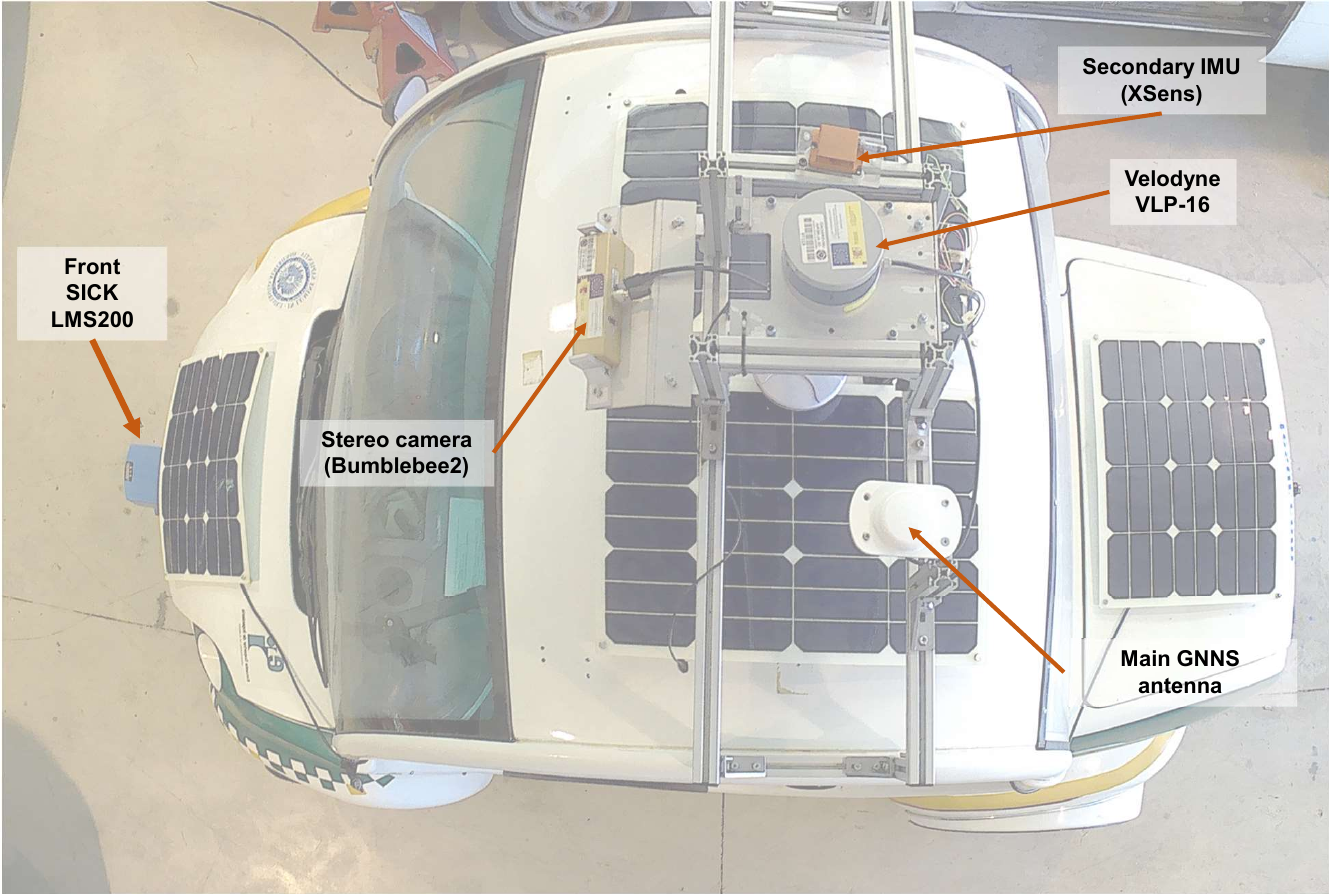}}
	\subfigure[]{ \includegraphics[width=0.49\columnwidth]{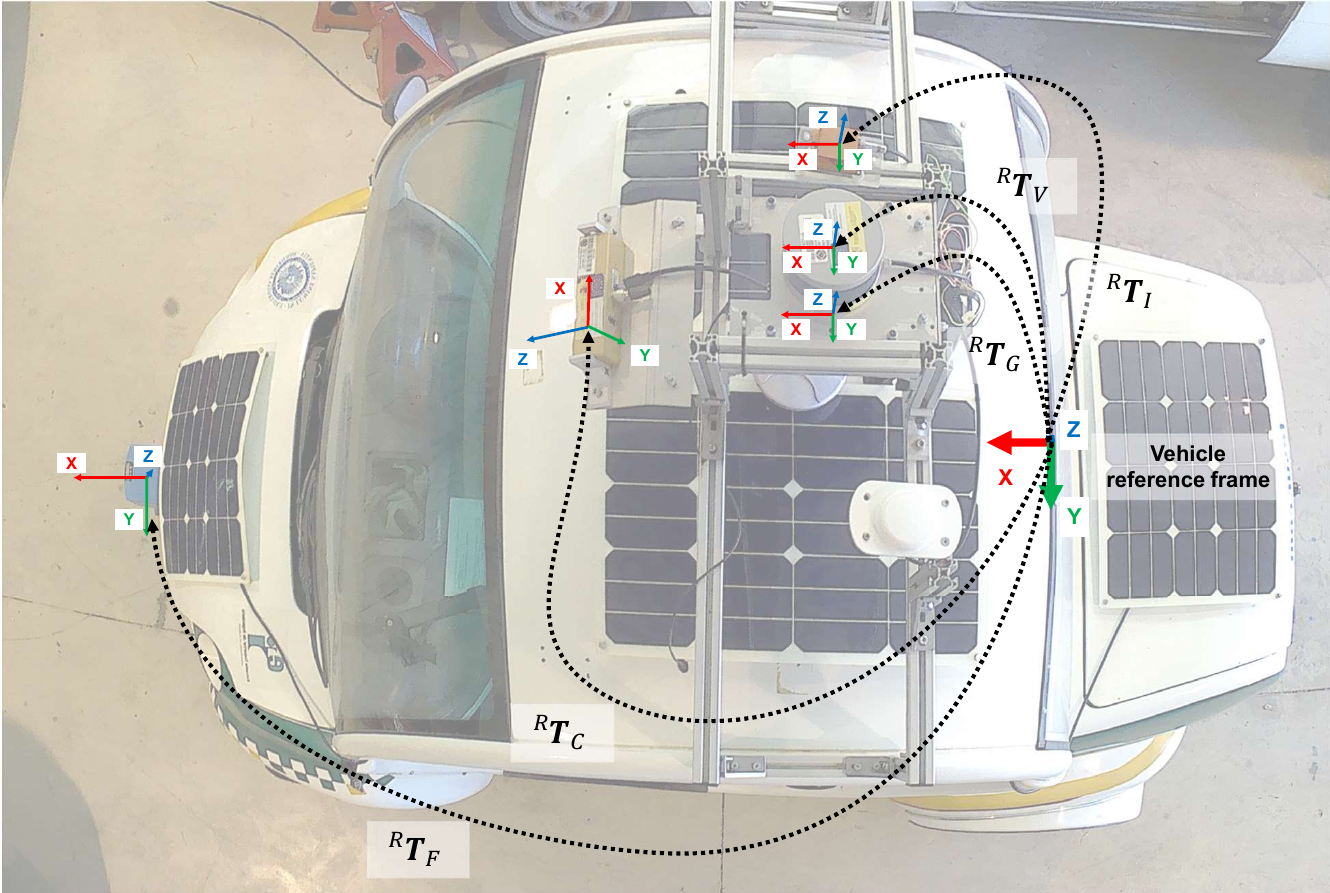}}
	\caption{Electric vehicle prototype used in the experimental~tests. (\textbf{a}) Side view: sensors; (\textbf{b})~Side~view: frames; (\textbf{c}) Top view: sensors; (\textbf{d}) Top view: frames}.
	\label{fig:car.frames}
\end{figure}

{   The main characteristics of the experimental vehicle used are summarized in Table~\ref{tab:ualCarCharacteristics}. A~pack of eight batteries Trojan TE35-Gel 210Ah 6V propels the vehicle ensuring an autonomy of 90 km at a maximum travel speed of 45~km/h by means of a 48 V DC motor controlled by a permanent magnet motor. Speed is controlled by a Curtis PMC controller (model 1268-5403). Three voltmeters are employed to measure the voltage in the rotor, the~field, and~the batteries. In~addition, the~prototype is equipped with three ampere-meters (LEM DHR 100, LEM, Fribourg, Switzerland) to measure instantaneous current consumption, at~the same three elements.}

{   Figure~\ref{fig:car.frames} shows all the installed sensors, together with their relative poses with respect to the vehicle frame of reference. Approximate values for each such poses, together with the raw dataset, are~available {online} ({\url{https://ingmec.ual.es/datasets/lidar3d-pf-benchmark/}).}


\end{document}